\documentclass[conference]{IEEEtran}
\IEEEoverridecommandlockouts
\usepackage{cite}
\usepackage{amsmath,amssymb,amsfonts}
\usepackage{graphicx}
\usepackage{textcomp}
\usepackage{xcolor}
\usepackage{fourier} 
\usepackage{array}
\usepackage{makecell}
\usepackage{tikz}
\usepackage{graphics}
\usepackage{epsfig}
\usepackage{subcaption}
\usepackage{amssymb}
\usepackage{amsfonts}
\usepackage{algorithm}
\usepackage{algpseudocode}
\usepackage{rotating}
\newcommand{\eatme}[1]{ }
\usepackage{multicol}
\usepackage{adjustbox}
\usepackage{subfloat}
\usepackage{float}
\usepackage{enumitem}
\usepackage{tablefootnote}

\usepackage{url}

\usepackage{breakurl}
\usepackage[breaklinks,hidelinks]{hyperref}

\def\BibTeX{{\rm B\kern-.05em{\sc i\kern-.025em b}\kern-.08em
    T\kern-.1667em\lower.7ex\hbox{E}\kern-.125emX}}
\begin{document}

\title{Video-based Pedestrian and Vehicle Traffic Analysis During Football Games
}

%\thanks{J.P.F., R.P., T.B., A.R., and S.R. are from the University of Florida, Gainesville, FL 32611. }

\author{Jacques P. Fleischer, Ryan Pallack, Ahan Mishra, Gustavo Riente de Andrade, Subhadipto Poddar,\\Emmanuel Posadas, Robert Schenck, Tania Banerjee, Anand Rangarajan, and Sanjay Ranka%
}

\maketitle

\begin{abstract}
This paper utilizes video analytics to study pedestrian and vehicle traffic behavior, focusing on analyzing traffic patterns during football gamedays. The University of Florida (UF) hosts six to seven home football games on Saturdays during the college football season, attracting significant pedestrian activity. Through video analytics, this study provides valuable insights into the impact of these events on traffic volumes and safety at intersections. Comparing pedestrian and vehicle activities on gamedays versus non-gamedays reveals differing patterns. For example, pedestrian volume substantially increases during gamedays, which is positively correlated with the probability of the away team winning. This correlation is likely because fans of the home team enjoy watching difficult games. Win probabilities as an early predictor of pedestrian volumes at intersections can be a tool to help traffic professionals anticipate traffic management needs. Pedestrian-to-vehicle (P2V) conflicts notably increase on gamedays, particularly a few hours before games start. Addressing this, a ``Barnes Dance'' movement phase within the intersection is recommended. Law enforcement presence during high-activity gamedays can help ensure pedestrian compliance and enhance safety. In contrast, we identified that vehicle-to-vehicle (V2V) conflicts generally do not increase on gamedays and may even decrease due to heightened driver caution.
\end{abstract}


\section{Introduction}

Traffic intersections serve as the most prominent hotspots for injuries and fatal crashes, as a typical four-legged intersection consists of 32 vehicle-to-vehicle conflict points and 24 vehicle-to-pedestrian conflict points~\cite{roundabouts}. The Federal Highway Administration states that each year, roughly $25\%$ of traffic-related fatalities and $50\%$ of traffic-related injuries occur at or near an intersection~\cite{isafety}. In addition to that, even as COVID-19 reduced traffic movement, pedestrian fatalities have gone up from 1,761 in 2017 to 1,788 in 2021~\cite{fars}. Thus, there is an eminent need for transportation professionals to identify strategies and mitigate such high-risk crash zones in the traffic realm.

One of the ways to address the issue is to use traffic cameras, as these can help apply artificial intelligence (AI) to analyze the traffic and pedestrian movements at the intersections. While transportation professionals keep on increasing traffic cameras at intersections~\cite{costs}, newer AI-based methods related to traffic safety have started evolving. Considering the evolution of traffic cameras and intersection-related safety studies, this paper focuses on using traffic cameras at an intersection to measure and track pedestrians and vehicles at signalized intersections. Tracking and quantifying their movements at intersections and road crossings can provide vital insights into pedestrian and vehicle behavior. One of the biggest issues with traffic cameras is the privacy concerns of the individuals and maintaining their anonymity~\cite{fries}. This study acknowledges privacy concerns and prioritizes the privacy and safety of road users by anonymizing any personal information captured by the cameras. %These data can serve as the backbone to model development that can aid transportation engineers, planners, urban developers, and policymakers in designing safer and more efficient road systems.

Moreover, beyond tracking and quantifying road users, the study further assesses potential risks and dangerous situations between pedestrians and vehicles. The research team uses %advanced artificial intelligence algorithms to identify and flag hazardous scenarios, such as 
new surrogate safety measures to quantify severe events \cite{banerjee}, which are determined by multiple existing metrics such as time-to-collision (TTC) and post-encroachment time (PET) as recorded in an event. Deceleration and speed characterize near-miss incidents or instances where pedestrians and vehicles come into proximity. To define proximity, previous studies have used a general range of around 1.5-3.0 seconds for TTC and 1.0-1.5 seconds for PET~\cite{banerjee}.

This critical analysis can help pinpoint high-risk areas and contribute to the development of targeted safety measures to reduce the likelihood of crashes.

Apart from day-to-day traffic safety concerns at intersections, special events like gameday pose a unique challenge to transportation professionals in terms of intense travel demand and high pedestrian flow~\cite{villiers}. Managing such event-based traffic using the aforementioned surrogate measures could be a viable option for transportation professionals. As such, this paper presents a case study of traffic safety and performance evaluation that can aid the traffic operations division in implementing traffic signal timing on a case-by-case basis to improve overall safety on such event-based days.

Overall, by responsibly leveraging these technological advancements, the research aims to enhance intersection safety, protect vulnerable road users, and ultimately contribute to the broader goal of creating safer and more sustainable urban environments.

In this paper, we offer an in-depth analysis of intersection safety utilizing video processing techniques along with their application for event-based days. The study contributes to the traffic analysis field in the following ways:
\begin{enumerate}
    \item Development of a specialized software tool aimed at aiding traffic engineers in analyzing traffic patterns during special events.
    \item Demonstration of the software's effectiveness through its application to gameday events at the University of Florida, which entails a crowd of over 90,000 people during the season.
    \item Utilization of the study's findings to devise appropriate countermeasures for enhancing the safety and performance of traffic intersections.
\end{enumerate}

The following sections outline the structure of our study. The 'Related Work' section reviews existing research on video processing for intersection safety analysis. Subsequently, we elaborate on our proposed methodology for intersection analysis. Next, we present our experimental findings and results, which were obtained from an intersection in close proximity to the event venue. Finally, drawing from our research, we present key conclusions and suggested countermeasures.

\section{Related Work}

Real-time object detection is a critical task in many computer vision projects, and significant research has been done regarding the most efficient algorithm to maintain accuracy and speed. Deep learning technologies, one being You Only Look Once version 4 (YOLOv4), have accomplished this task better than any physical sensor or detector. Bochkovskiy et al. \cite{yolov4} establishes the effectiveness of the YOLOv4 algorithm, which is also used in this paper, for real-time object detection. Our approach leverages YOLOv4 via fisheye cameras strategically positioned at intersections to enable real-time analysis of vehicle and pedestrian safety.

While the previous work focused on the vehicle detection capabilities of YOLOv4, another paper by Kathuria and Vedagiri explored how pedestrian safety interacts with vehicles by using a proactive approach to evaluate pedestrian-vehicle interaction dynamics at non-signalized intersections. They specifically used video recordings and traffic simulations to accomplish such an approach~\cite{kathuria}. Some of the restrictions of this work lie in the form of a short duration of analysis (450 minutes), limited conflicts involving pedestrians, and the absence of appropriate visualizations like heatmaps. %This paper analyzes pedestrian safety, however, there are key differences. The study utilizes four non-signalized intersections from Navi Mumbai or Nagpur, both Indian cities, and analyzes only 450 minutes of footage. Specific macro information about the intersection, such as heatmaps to display the changing volumes of pedestrians and conflicts by hour, is not gathered in this study. Further, with the four intersections being vastly different in vehicle and pedestrian volumes, categorizing of conflicts based on location is instead replaced with two patterns, where it is pattern 1 if either pedestrian or vehicle or both take an evasive action, else it is pattern 2.
	
 Once we have established the feasibility of analyzing intersections using real-time video analytics, the next step involves utilizing important safety metrics such as TTC and PET to quantify and assess the safety performance of these intersections.
TTC refers to the remaining time a road user has to avoid a collision, starting from the moment they take evasive action until the potential point of collision~\cite{banerjee}. On the other hand, PET represents the time difference between the departure of the first road user from a specific point and the subsequent arrival of the second road user at the same point~\cite{banerjee}. Other work implementing TTC and/or PET, such as papers by Olszewski~\cite{olszewski} and Warchol~\cite{warchol}, contains oversights such as not automatically calculating PET as well as being able to use both statistics rather than only PET.

Along with these metrics, the classification of P2V conflicts from Type 1 to Type 6 allows us to further understand the nature and severity of different interactions between road users and pedestrians. By categorizing conflicts based on their specific types, such as right-turning vehicles with pedestrians on adjacent parallel crosswalks (Type 1) or left-turning vehicles with pedestrians on far-side crosswalks (Type 3), we can gain valuable insights into the most common conflict scenarios and identify patterns that require targeted safety improvements at intersections. Further, Mishra et al.~\cite{mishra} and Banerjee et al.~\cite{banerjee} also discuss the idea of metrics and conflict classification. This paper leverages these concepts to help understand how a gameday, which is a football game happening in the stadium on the college campus, affects pedestrian and vehicle safety.

Moreover, a study by Austin et al. has described an increase in dangerous traffic situations on football gamedays~\cite{austin}. Another study by Xiong et al. focused on traffic surrounding the 2010 Asian Games describes an Internet-of-things system that dynamically changes traffic based on observed vehicle trajectories, specifically for supporting a sporting event~\cite{xiong}. The study focuses on optimizing traffic flows by ingesting significant traffic information. Additionally, the study takes a holistic approach to traffic analysis, considering a network of roads and routes rather than a specific intersection with detailed analysis. However, the studies could be improved by specifically looking at pedestrian interactions within the intersections so as to minimize the number of dangerous situations they encounter.

Aside from establishing the trend that nearby events (and, likewise, more pedestrians) can cause more conflicts, papers within the transportation research field have also sought to find mitigation measures. Kattan et al. identified that a pedestrian scramble operation (PSO) (also called ``Barnes Dance'') significantly decreased the number of P2V conflicts within the city of Calgary, Alberta, Canada~\cite{kattan}. While we do not appraise PSO efficacy within this paper, we consider how such measures may affect the volume of conflicts.
\begin{figure*}[h]
\centering
\includegraphics[width=2\columnwidth]{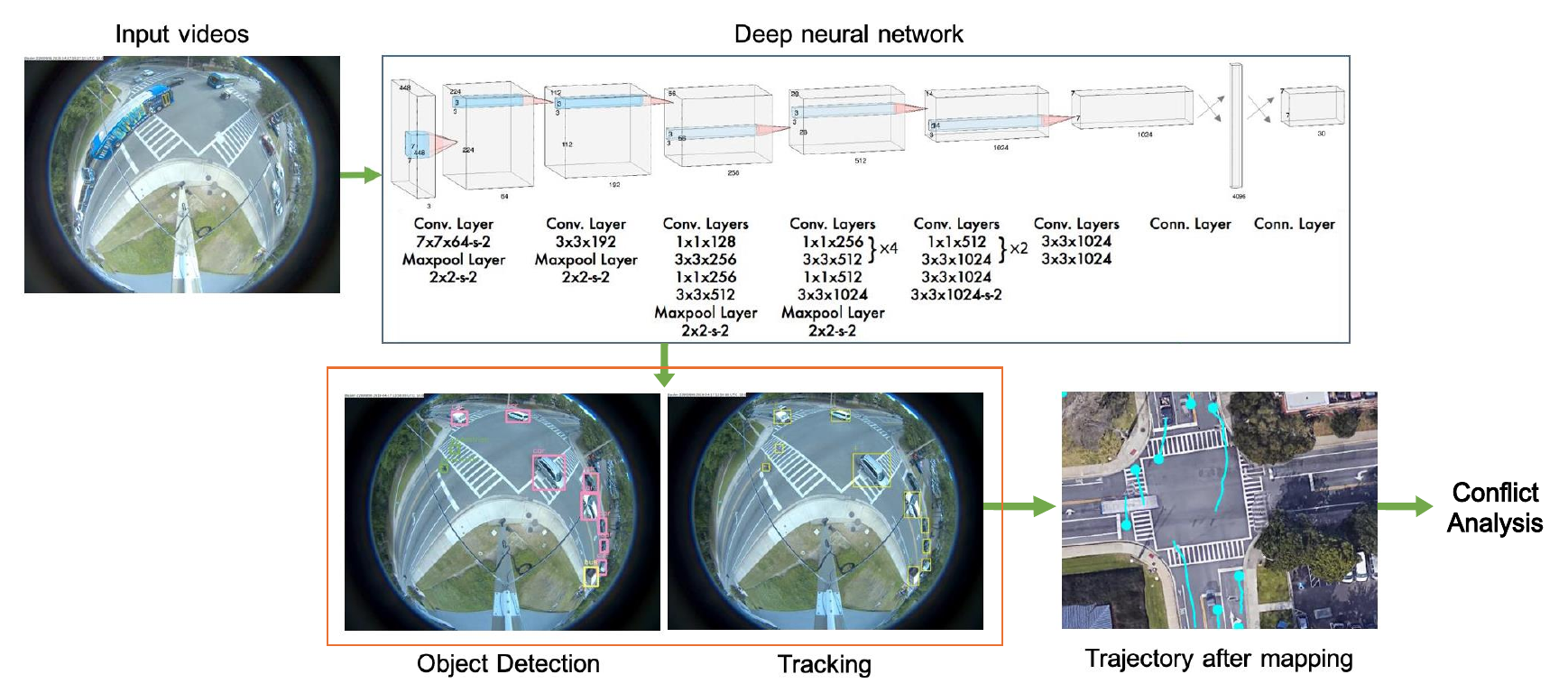}
% \vspace{-1.5cm}
\caption{The pipeline and deep model architecture used in this paper that has been adapted from our previous work~\cite{huang}.}
\label{fig:pipe}
\end{figure*}

The findings of these papers have shown the need for traffic analysis, as well as YOLOv4's potency for real-time object detection, while also providing insights into pedestrian safety and interaction dynamics at intersections. However, the existing literature lacks a comprehensive analysis of pedestrian-vehicle interactions, especially during sporting events with a surge of pedestrians every weekend in colleges all around the country. Our research aims to bridge this gap by employing YOLOv4 with fisheye cameras to study pedestrian behavior, conflicts, and safety patterns at intersections during game-days. By examining macro traffic information and categorizing conflicts based on different phases and movements, our study intends to contribute new knowledge to the field and offer valuable data for enhancing traffic management strategies to prioritize both vehicle and pedestrian safety, especially during the short surge due to sporting events.

\section{Methods}
Modern advancements in video technology, particularly fisheye cameras, have enabled detailed analysis of intersections, providing valuable data to enhance road safety. This paper presents an innovative video analytics system that efficiently processes intersection videos, extracts vital trajectory data, and explores various safety aspects by investigating two types of conflict events: pedestrian-to-vehicle (P2V) and vehicle-to-vehicle (V2V). The study systematically evaluates pedestrian and traffic volumes, enabling the identification of peak periods and potential areas of concern. Combining volume studies with safety analysis ensures effective countermeasure proposals that minimize disruptions to overall traffic patterns.
\subsection{Video Processing}
Figure~\ref{fig:pipe} illustrates our video processing pipeline, which has been adapted from our previous work~\cite{huang}. The video analysis software processes fisheye videos as input, employing various techniques to extract object information from each video frame. First, it draws bounding boxes around identified objects within the frame, determining their respective classes based on previously annotated images. The software then derives the (x, y) coordinates of each object, considering the center point of its bounding box.

The fisheye cameras installed at intersections capture ten video frames per second, providing a static snapshot of the traffic scenario. The object detection and tracking module utilizes the YOLOv4 algorithm~\cite{yolov4} to detect different road participants, such as cars, buses, trucks, pedestrians, and motorcyclists.

For streamlined object tracking across frames, a modified DeepSORT algorithm associates object detections and assigns a unique ID to each object throughout the video. However, a post-processing step is necessary since the (x, y) coordinates are initially in the circular fisheye video space. This step involves transforming the coordinates into rectilinear space using fisheye-to-perspective transformation, followed by thin-plate spline (TPS) warping. As a result, the software generates new timestamped coordinates in the rectilinear space, providing a more usable representation of object locations.

The software stores the timestamped coordinates in a dedicated database (DB) table to efficiently manage the processed data, facilitating further analysis and studies of intersection activity and traffic patterns. This approach ensures that privacy and safety concerns are thoroughly addressed throughout the video processing steps, aligning with ethical guidelines and data protection regulations.

\subsection{Conflict Analysis} %for Tania to fill in
Our software calculates two essential metrics: time-to-collision (TTC) and post-encroachment time (PET). Below, we offer a concise overview of the algorithm for each metric.
\subsubsection{Time-To-Collision (TTC)} The TTC computation involves examining all pairwise interactions among objects present at the intersection. Consequently, if there are $n$ objects at the intersection, there can potentially be $n \times (n-1)$ interactions. However, due to the large number of users at any given intersection within a time window, the quadratic expression $n^2$ can lead to computational inefficiencies. To address this issue, we implement commonsense filters that effectively reduce the search space. These filters, elaborated in detail in \cite{banerjee}, encompass specific observations to identify severe P2V or V2V conflicts. For V2V conflicts, for example, the users involved must be on conflicting trajectories, located within a distance of 10 meters from each other, and engaged in conflicting paths over multiple timestamps. This process is designed to exclude chance encounters between objects that might cross paths incidentally.

Once the filters have been applied, we analyze the remaining interactions, considering factors such as the location of the two objects, the distance between them, and their current speed values. The various cases considered for analysis are described further.

\paragraph{Case 1: Stationary Collision}

\begin{figure}[h!]
\centering
\includegraphics[clip, trim=0.5cm 2cm 0.5cm 3cm, width=1\columnwidth]{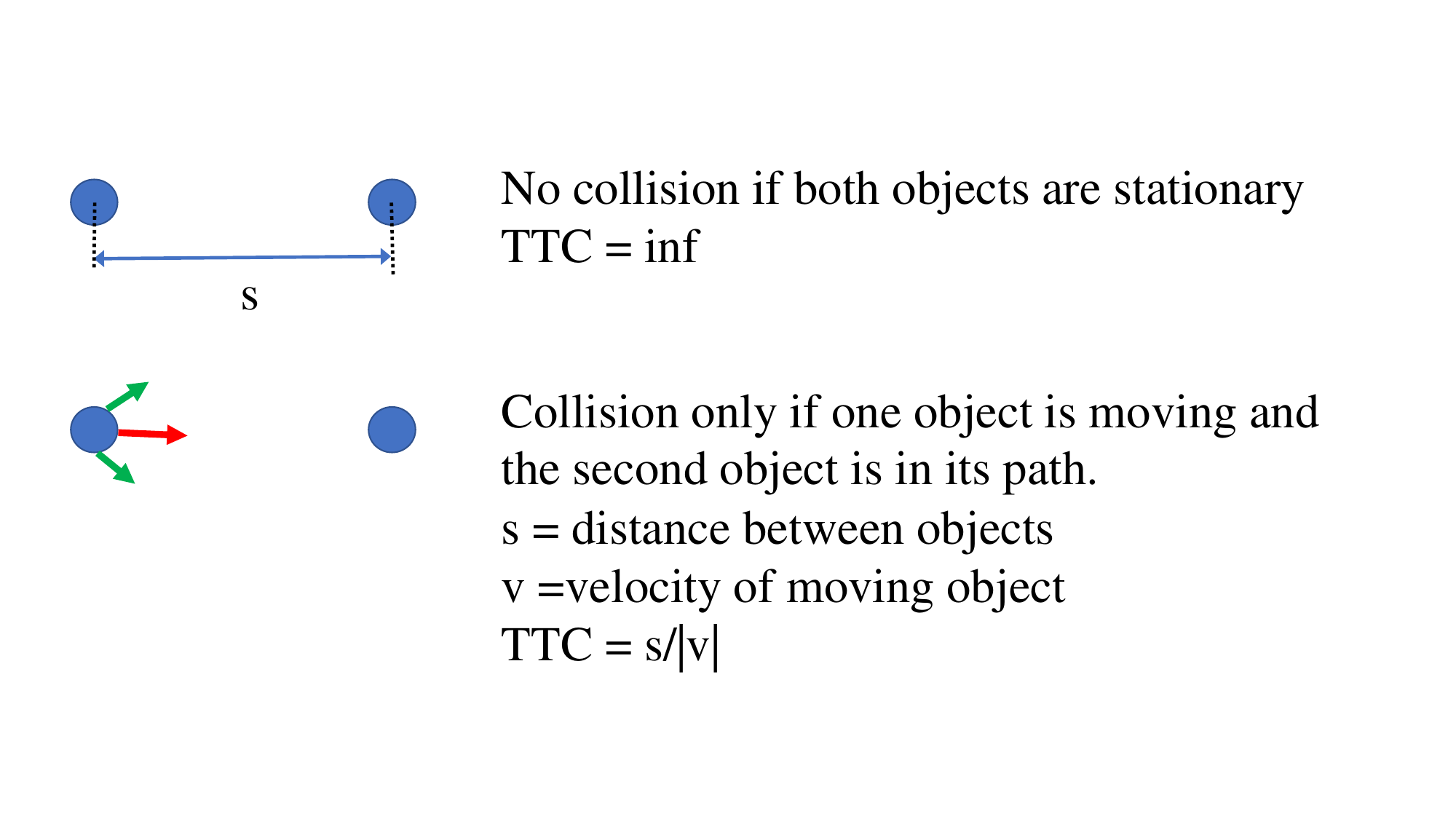}
% \vspace{-0.5cm}
\caption{Case 1 in TTC computation involves scenarios where either one or both objects are stationary, or one of them is in motion. When at least one object is moving, a potential conflict arises only if the path of movement aligns with the line connecting the two objects.}
\label{fig:ttc1}
\end{figure}

Figure~\ref{fig:ttc1} depicts a scenario that involves one or both users being stationary. When both users are stationary, there is no possibility of a collision, and the TTC is considered infinite. However, if one of the objects is in motion while the other remains stationary, a potential collision can occur, but only if the stationary user lies in the path of movement of the other user.

The TTC for this specific case can be calculated as shown in Equation~\ref{Eqn:TTC1},
\begin{equation}\label{Eqn:TTC1}
t = \frac{s}{|v|}
\end{equation}
where $|v|$ is the speed (magnitude of velocity $v$), and $s$ is the distance between the objects. 

The TTC is determined for each frame captured during video processing, typically at intervals of 0.1 seconds. Consequently, any acceleration or deceleration applied by the user is automatically accounted for by observing the difference in user speed from one frame to another. During video processing, we obtain the velocity components along the horizontal and vertical axes to aid in this computation.

\paragraph{Case 2: Passing Collision}

\begin{figure}[h]
\centering
%\vspace{-2cm}
\includegraphics[clip, trim=0.5cm 3cm 0.5cm 4cm,width=1\columnwidth]{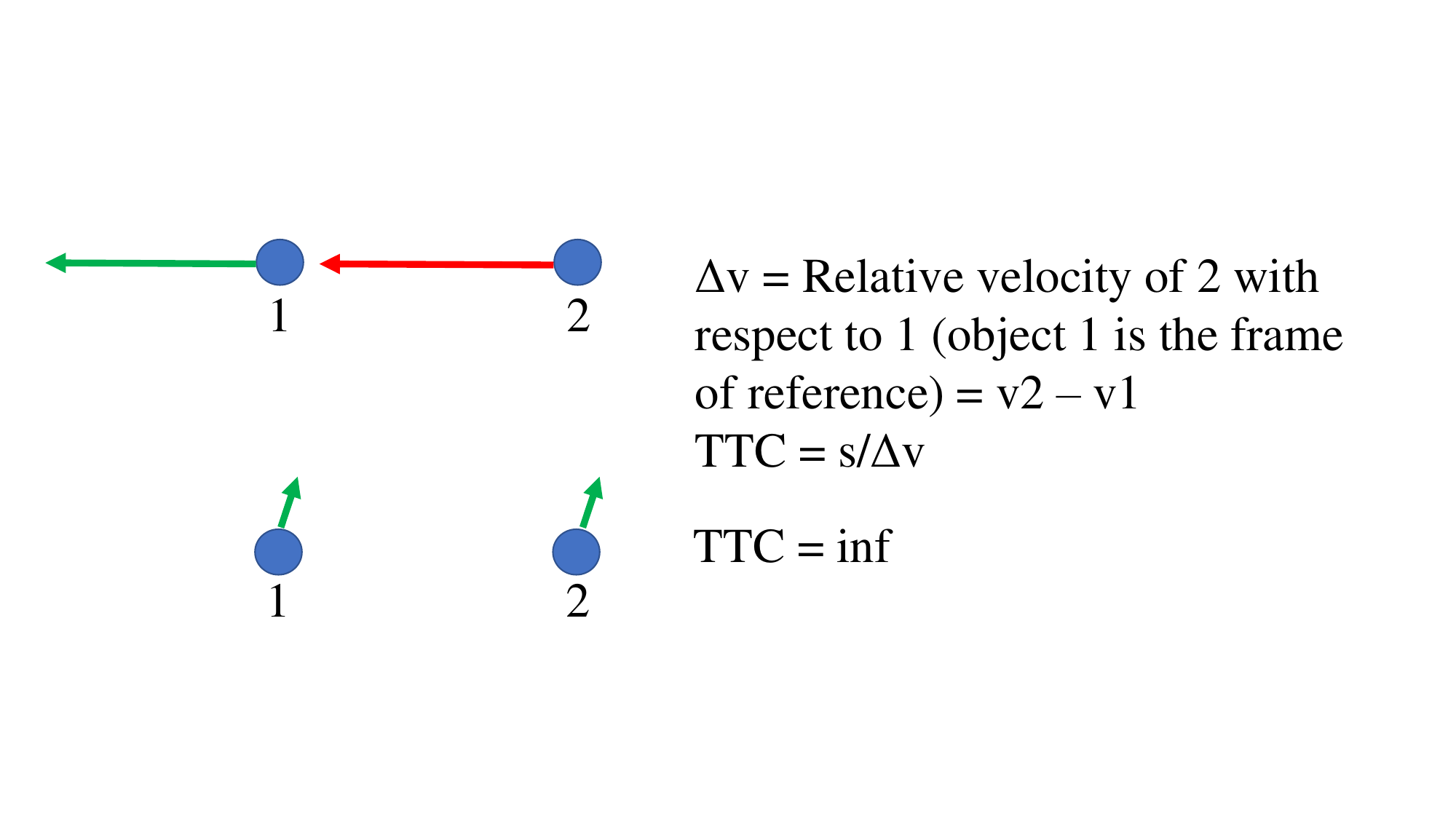}
% \vspace{-0.5cm}
\caption{Case 2 of TTC computation involves scenarios where both objects are moving along parallel lines. In this case, a collision is only possible when the lines are coincident, meaning they are directly on the same path, in which case TTC is calculated using the formula: TTC = distance between the objects / relative velocity of the objects.}
\label{fig:ttc2}
\end{figure}

As depicted in Figure~\ref{fig:ttc2}, when two entities are moving along parallel lines, a collision can only occur when they are on the same path. All other cases generally correspond to scenarios where entities are moving on different lanes; the TTC is regarded as infinity. When the entities are moving on the same line, we proceed to compute their relative velocity. The TTC for the two objects is obtained by dividing the distance between them by their relative velocity. The calculation formula is represented as Equation~\ref{Eqn:TTC_2},
\begin{equation}\label{Eqn:TTC_2}
t = \frac{s}{\Delta v}
\end{equation}
where $\Delta v$ is the relative velocity while $v_2$ and $v_1$ are the velocities of the following and leading objects, respectively. 

Let $|v_2|$ and $|v_1|$ represent the speed of the objects. If $|v_2| < |v_1|$, the relative velocity is negative, resulting in a negative TTC value, indicating that there is no chance for a collision. Conversely, if $|v_2| > |v_1|$, a positive and viable TTC value is obtained, suggesting a potential collision scenario. However, if $|v_2| = |v_1|$, the TTC becomes infinity,  indicating again that there is no possibility of a collision in this situation.

\paragraph{Case 3: Head-On Collision}

\begin{figure}[htp]
\centering
\includegraphics[clip, trim=0.5cm 3cm 0.5cm 4cm, width=1\columnwidth]{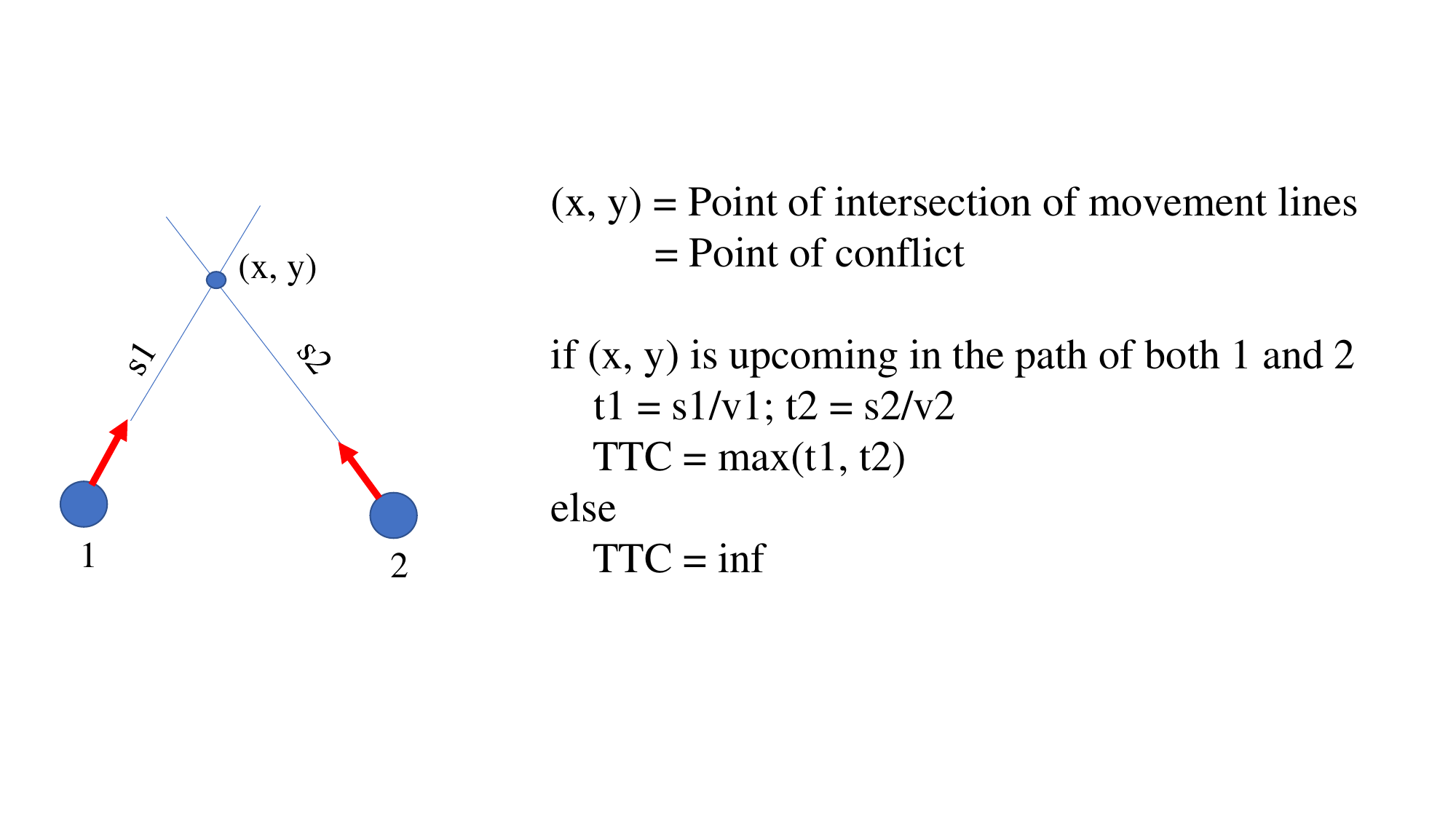}
% \vspace{-0.5cm}
\caption{Case 3 is the most general case of conflict occurrence, which involves two objects approaching each other from different directions. The conflict point, where their lines of motion intersect, is first determined. The distance of the objects 1 and 2 from the conflict point are represented as s1 and s2, respectively.}
\label{fig:ttc3}
\end{figure}

As illustrated in Figure~\ref{fig:ttc3}, this is the most general case of conflict occurrence; it could happen in the middle of the intersection where entities from different directions cross the intersection. These conflicts would be extremely rare if everyone followed the signaling and yielding rules. For Case 3, we first find the point of intersection of the lines of motion of the two entities. This point of intersection is called the conflict point. Once determined, we compute the time for the first and second entities to reach that conflict point. If it takes the same time for both entities to reach the conflict point, then that time is the TTC. Otherwise, we consider TTC the maximum time required for the entities to reach the conflict point.

If the first entity is expected to completely pass the conflict point before the second entity reaches it, then there is no possibility of a collision. This happens, for instance, in a correctly and uniformly followed transition of traffic sequences; eventually, as one phase features crossing the intersection, a subsequent phase will perpendicularly cross the previous phase's paths. However, there is no chance of collision since such time has passed between phases for this to occur safely. Furthermore, the filter does not account for such usual scenarios due to the previously mentioned commonsense filters.

\subsubsection{Post-Encroachment Time (PET)}
\begin{figure}[h]
\centering
\includegraphics[width=1\columnwidth]{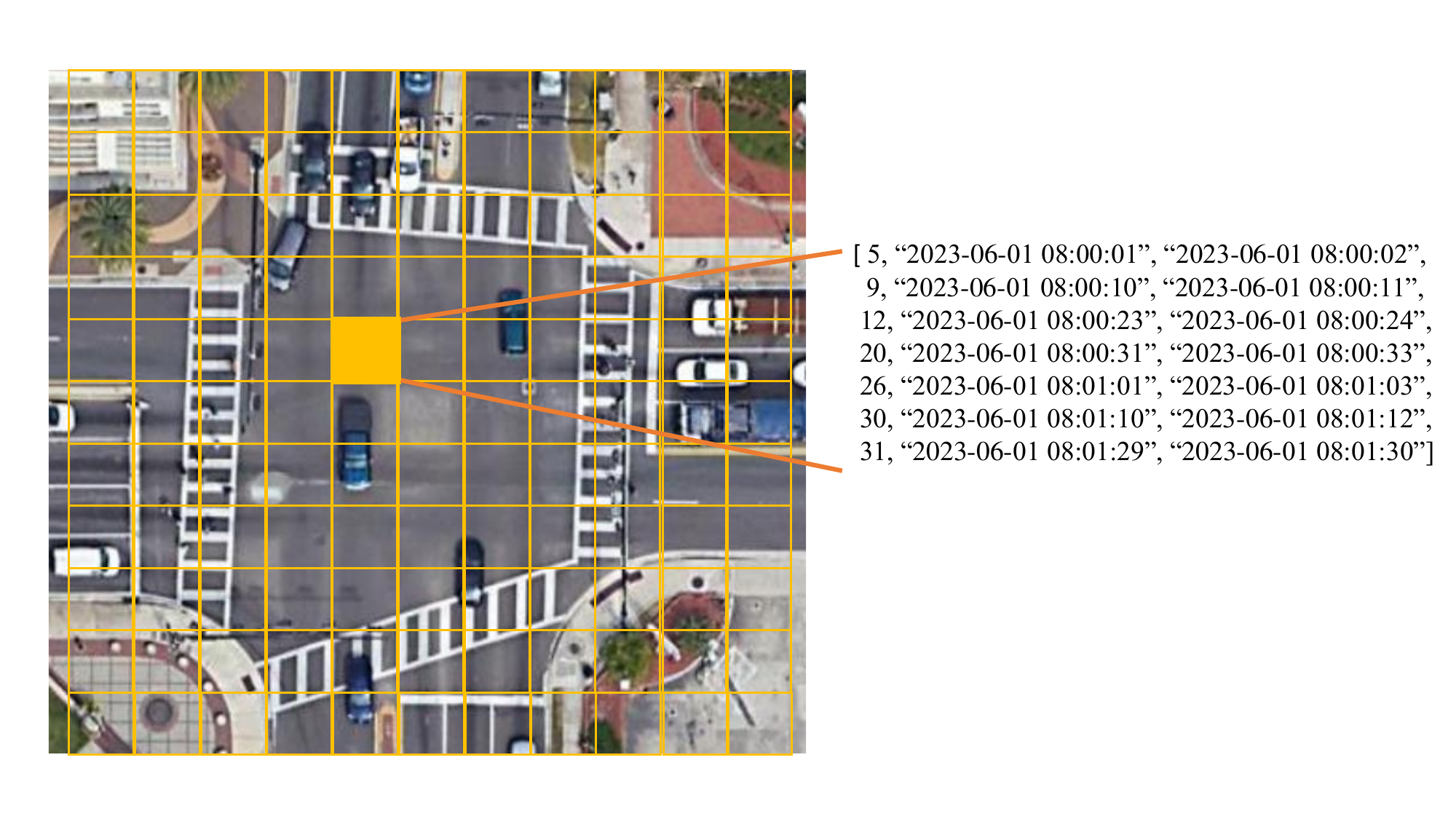}
\caption{Overview of PET computation process, involving mesh division of the intersection region, and object list creation. PET events are identified when objects interact and move over the same region within a short period, potentially indicating significant events.}
\label{fig:pet}
\end{figure}

PET computation involves several steps. First, the intersection region is divided into a mesh grid, and for each cell in the grid, a list of objects that went over that cell (from any direction) is created. This list includes the object ID and the timestamps when the object entered and left the cell. Figure~\ref{fig:pet} illustrates the overview of the PET computation process involving mesh division of the intersection region and object list creation.

Next, the algorithm iterates through all cells with non-empty lists and identifies consecutive objects in the list. For each pair of consecutive objects, the algorithm calculates the time difference between when the leading object left the cell and the following object entered the cell. If the time difference falls within a certain threshold, such as $10$ seconds in our software, the event is recorded as a conflict. This means that the objects interacted within a short period, indicating a potentially significant event. For a significant event (we refer to such as a severe event), the PET is around 2 seconds for TTC and 3 seconds for PET~\cite[p. 76]{tania-book}. 

In summary, PET computation involves mesh division, object list creation, time difference calculation, and event recording based on a specified threshold.

This section described the algorithms for determining TTC and PET. TTC calculation involves intricate considerations of user trajectories, relative velocities, and distance measurements. By identifying critical cases of interaction, the TTC algorithm provides essential insights for intersection safety. On the other hand, PET computation utilizes mesh division and time difference analysis to detect interactions between objects. This information proves invaluable in understanding noteworthy events that occur within short time frames. The integrated algorithms for TTC and PET computation, coupled with the analysis of deceleration, speed, and inter-object distance, play a pivotal role in identifying severe events. 

\subsection{Data Retrieval and Preprocessing}
%To conduct the analysis, the data retrieval and preprocessing steps involve querying the MySQL database to retrieve trajectories specific to a particular intersection. The database returns the entries as a CSV file, which is then imported into a Jupyter notebook. Utilizing the pandas Python module, the data is processed to ensure there are no duplicate trajectories for any particular object, focusing only on anomalous entries.

To conduct the analysis, the data retrieval and preprocessing steps involve querying the database to retrieve trajectories for an intersection. The retrieved trajectories were grouped according to the time-of-day and day-of-week based on the different signal timing patterns used for this case study. The heatmaps for various conflicting diagrams were then prepared accordingly. 

%In the preprocessing phase, timestamps within the trajectories are converted to Pythonic datetime objects, and a new "day-of-the-week" column is created to facilitate further analysis. Moreover, each timestamp is categorized into specific time-of-day segments, as illustrated in Table~\ref{tab:times}.

%Once the trajectories are organized by dates, they are further separated based on the hour digit. Subsequently, these trajectories are visually represented on heatmaps, with the y-axis denoting the phase and the x-axis representing the hour. The intensity of the color on the heatmap indicates the number of identified pedestrians walking during specific time intervals.

%Additionally, another type of heatmap is generated to illustrate the occurrences of pedestrian-to-vehicle (P2V) conflicts across various times-of-day. This comprehensive visualization aids in understanding the patterns and trends associated with pedestrian and vehicle interactions.

\subsection{Phases}

Trajectories are classified into distinct numerical phases, wherein pedestrian-type phases are represented by even numbers ranging from 2 to 8, inclusive, corresponding to the appropriate vehicular phases. Additionally, Phase 0 is introduced to indicate anomalous behavior when the model is unable to categorize the movement to any particular phase described before. Such anomalous behavior could involve jaywalking scenarios like pedestrians traveling along medians between lanes that move in opposite directions.

\begin{figure}[h]
\centering
% shift it to the left
\hspace*{-0.15\columnwidth}
% cut the remnants at the top of the image with trim
\includegraphics[width=1\columnwidth, trim=0 0 0 5, clip]{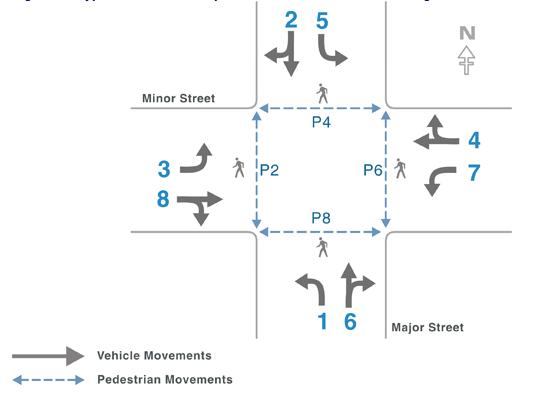}
\caption{The different possible phases for pedestrian and vehicle movement.}
\label{fig:phases}
\end{figure}

Figure~\ref{fig:phases} illustrates the various possible phases for pedestrian and vehicle movement. Specifically, Phases 2 and 6 represent north-south movements (major road), while Phases 4 and 8 depict east-west movements (minor road) in this figure.

\subsection{Conflict Type Classification}

\begin{figure}[h!]
    \centering
    \includegraphics[width=0.6\columnwidth]{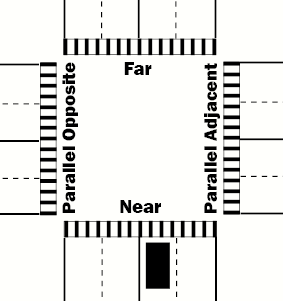}
    \caption{Crosswalk naming convention with respect to the vehicle, adapted from~\cite{tania-book}.}
    \label{fig:naming-convention}
\end{figure}

\begin{figure}[h!]
  \captionsetup[subfigure]{justification=centering}
  \centering
  \begin{subfigure}{0.2\textwidth}
    \includegraphics[width=\linewidth]{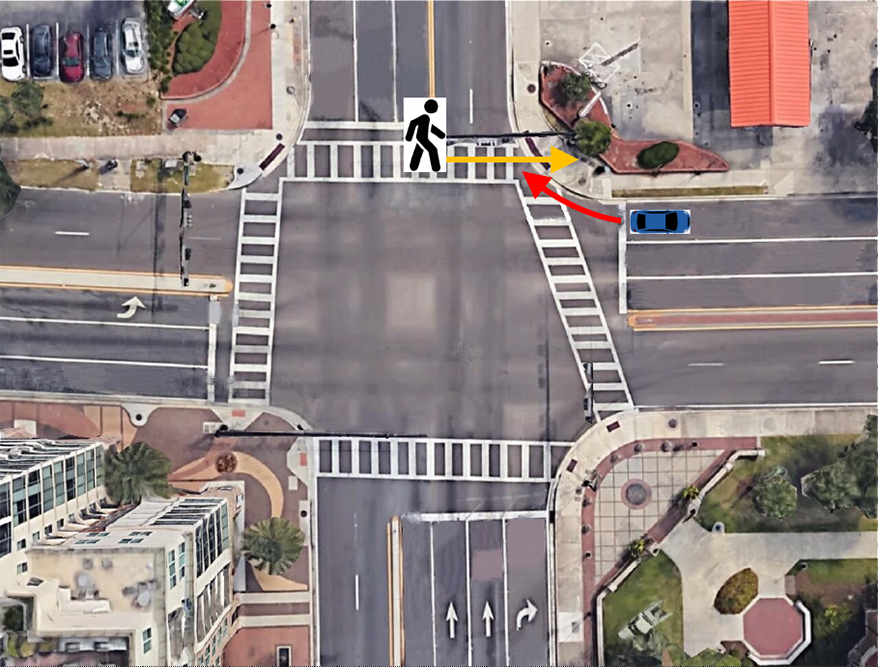}
    \caption{P2V: Type 1}
    \label{fig:conflict-types:a}
    \centering
  \end{subfigure}
  % \hspace{0.05\textwidth} % Adjust the horizontal space here
  \begin{subfigure}{0.2\textwidth}
    \includegraphics[width=\linewidth]{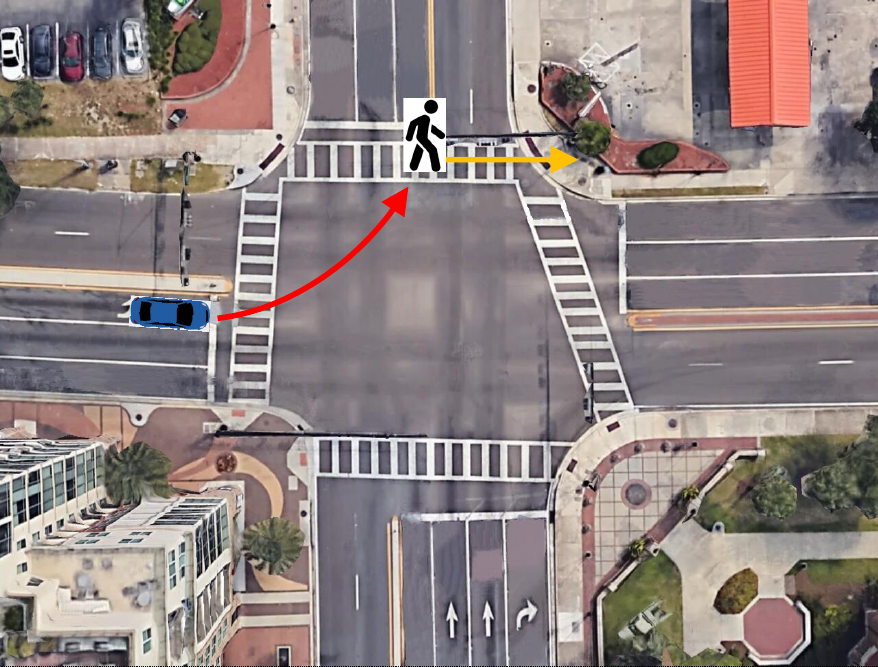}
    \caption{P2V: Type 3}
    \label{fig:conflict-types:b}
    \centering
  \end{subfigure}

  % \vspace{1\baselineskip} % Adjust the vertical space here

  \begin{subfigure}{0.2\textwidth}
    \includegraphics[width=\linewidth]{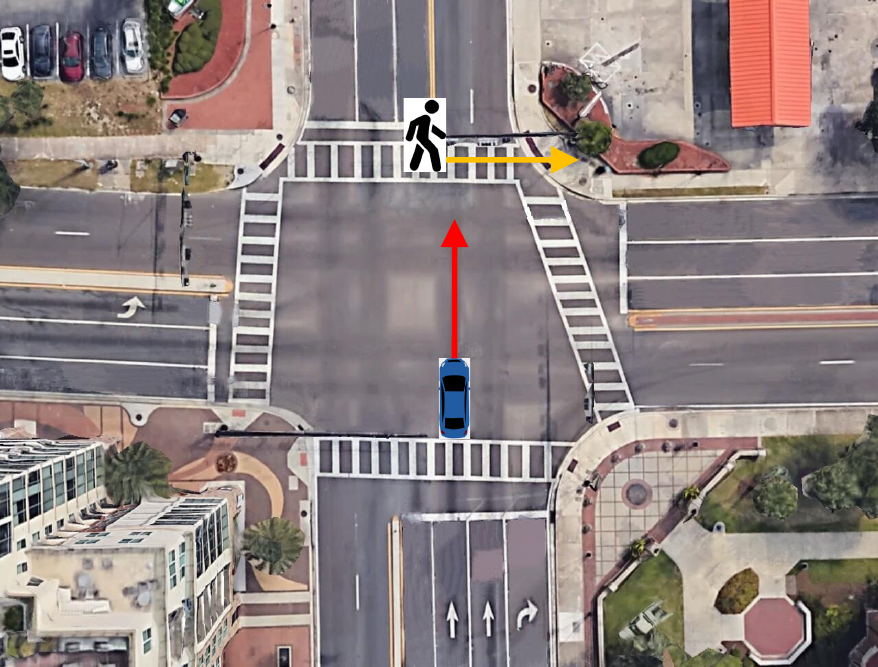}
    \caption{P2V: Type 5}
    \label{fig:conflict-types:c}
    \centering
  \end{subfigure}
  % \hspace{0.05\textwidth} % Adjust the horizontal space here
  \begin{subfigure}{0.2\textwidth}
    \includegraphics[width=\linewidth]{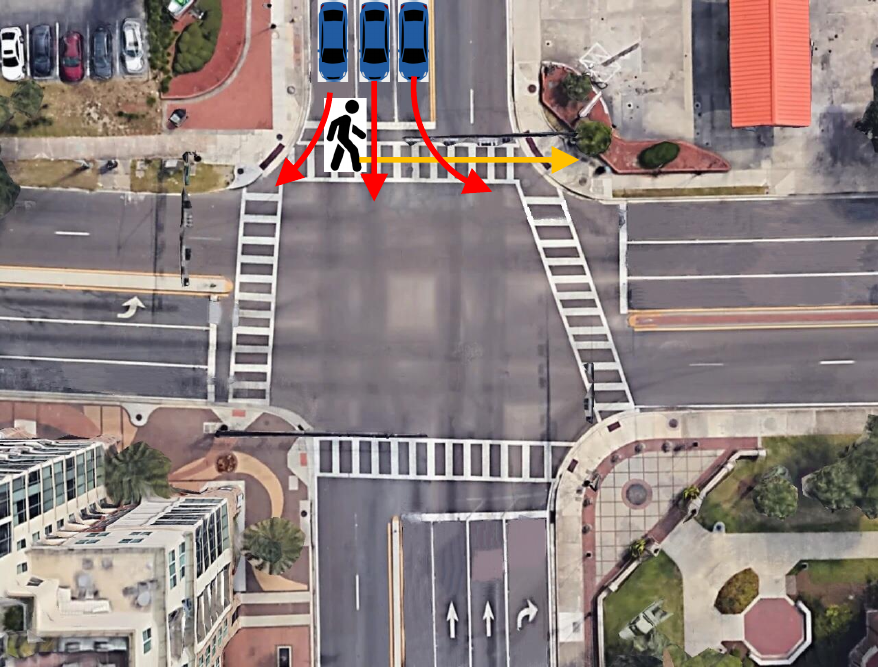}
    \caption{P2V: Type 2, 4, 6}
    \label{fig:conflict-types:d}
    \centering
  \end{subfigure}
  \caption{P2V conflict types, adapted from~\cite{tania-book}.}
  \label{fig:conflict-types}
\end{figure}

For specificity regarding conflict circumstances, our data pipeline classifies the P2V conflicts depending on the entities' trajectory and positioning within the intersection. To classify the P2V conflicts, we utilize a naming convention described in Figure~\ref{fig:naming-convention} and a numbering system shown in Figure~\ref{fig:conflict-types}.

In further detail, 

\begin{enumerate}
\item Conflict Types 1 and 2 are shown in Figures~\ref{fig:conflict-types:a} and \ref{fig:conflict-types:d}, involving a right-turning vehicle with a pedestrian in the adjacent parallel crosswalk and the near crosswalk, respectively. 
\item Conflict Types 3 and 4 are shown in Figures~\ref{fig:conflict-types:b} and \ref{fig:conflict-types:d}, involving a left-turning vehicle with a pedestrian in the parallel opposite crosswalk and the adjacent parallel crosswalk, respectively.
\item Conflict Types 5 and 6 are shown in Figures~\ref{fig:conflict-types:c} and \ref{fig:conflict-types:d}, involving a through vehicle with a pedestrian in the far crosswalk and the near crosswalk, respectively.
\end{enumerate}

Within our analysis, we utilize this classification to determine the manner and circumstances surrounding conflicts.

\section{Results}

This section focuses on presenting the intersection of interest, providing a detailed description of the gamedays, and conducting an analysis of vehicle and pedestrian volume along with conflict data. 
\subsection{Intersection Description}
%- Location of the intersection
%- Geometry of the intersection
%- Camera placement
%- Results, or comments about Normal traffic at this intersection
%- Table 1 in the methodology section should come here.
The chosen intersection of University Ave at 13th Street, located in the city of Gainesville, Florida, represents a standard 8-phase intersection serving vehicles and pedestrians. The intersection is used to examine various traffic dynamics and pedestrian movements.

\begin{figure*}[ht!]
\centering
\begin{subfigure}[t]{0.6\textwidth}
    \centering
    \includegraphics[width=\columnwidth]{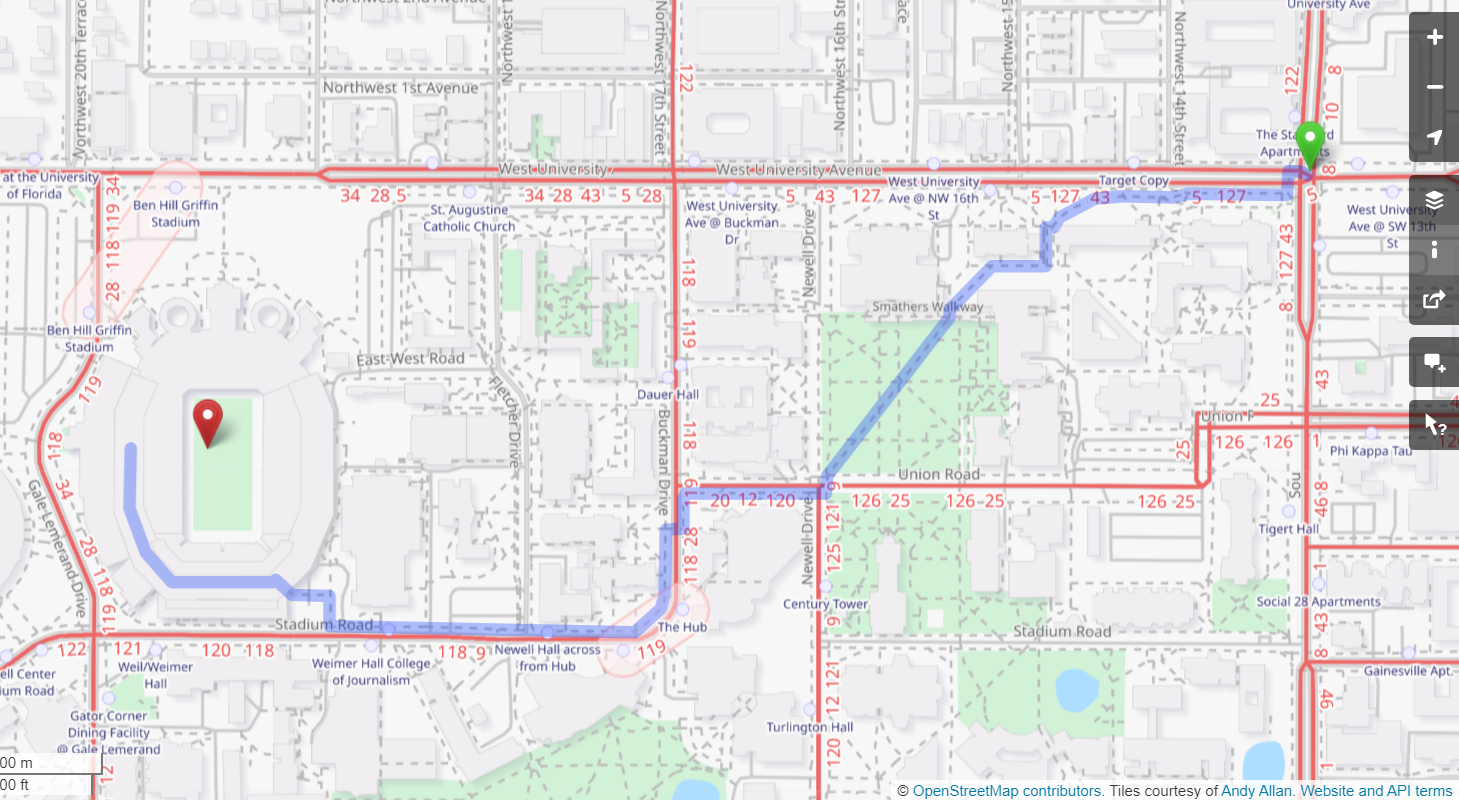}
    \caption{A pedestrian path from University Ave \& 13th St. (green) to the stadium (red) created using OpenStreetMap, which takes 15 minutes to walk~\cite{OpenStreetMap}.}
    \label{fig:pedestrian-path-map}
\end{subfigure}
~
\begin{subfigure}[t]{0.26\textwidth}
\hspace{20pt}
    \centering
    \begin{tikzpicture}[overlay, remember picture]
        \node[inner sep=0pt] (right_image) at (0,0) {\raisebox{170pt}{\includegraphics[width=\textwidth]{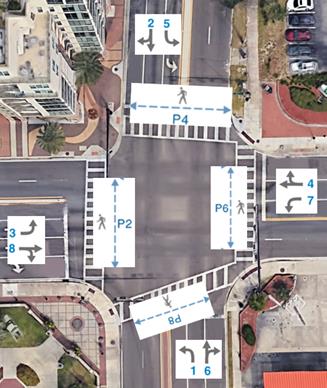}}};
        \draw[->] (right_image.north west) ++(0cm, -0.04cm) -- ++(-1.6cm, -0.6cm); % the last two are the , higher# = higherpos
        \draw[->] (right_image.north west) ++(0cm, -5.57cm) -- ++(-1.7cm, 4.2cm);
    \end{tikzpicture}
    
    \caption{Zoomed-in view of the intersection indicated in the green marker in (a).}
    \label{fig:univ-13-inter}
\end{subfigure}

\begin{subfigure}[t]{0.1\textwidth}
    \centering
    \includegraphics[width=\columnwidth]{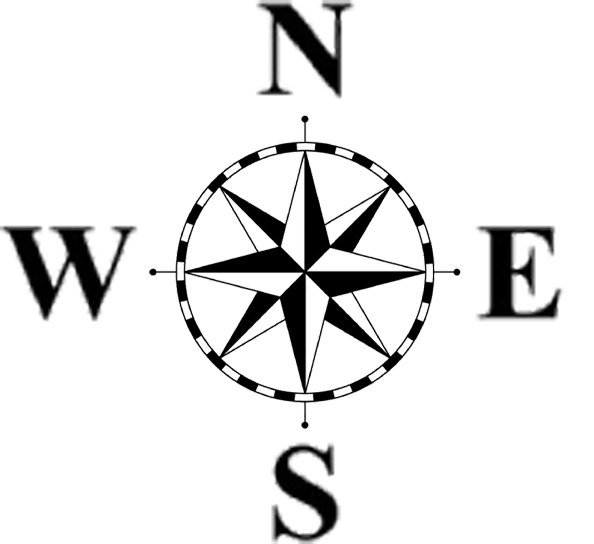}
\end{subfigure}
\caption{A geographical view of the intersection analysis.}
\label{fig:univ-ave-13-st}
\end{figure*}

\begin{figure}[h!]
\centering

\begin{subfigure}[h!]{.5\textwidth} % Adjust the width based on your preference
    \centering
    \includegraphics[width=0.9\textwidth]{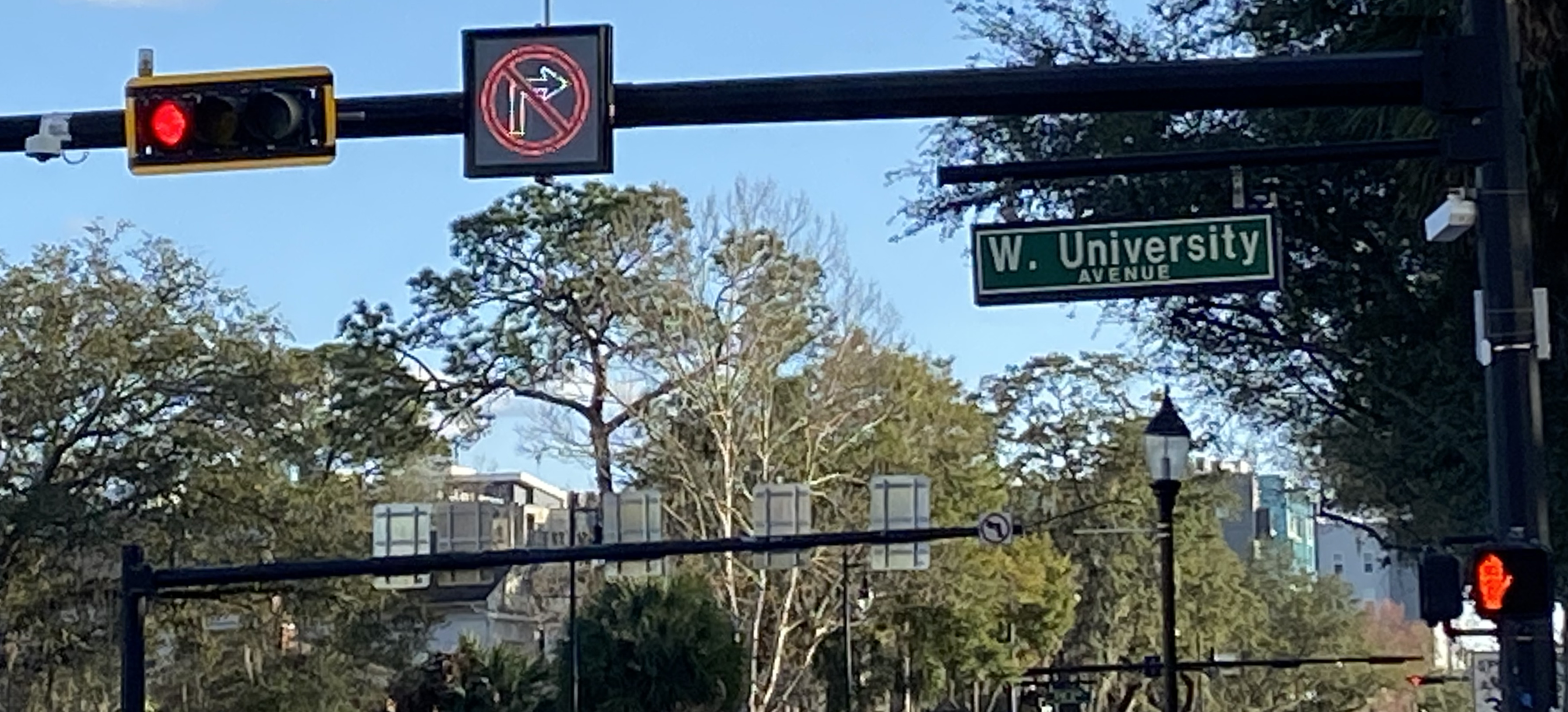}
    \captionsetup{justification=justified, singlelinecheck=false} % Adjust the justification here
    \caption{}
    \label{fig:no-right-turn-sign}
\end{subfigure}

\vspace{0.5cm} % Adjust the value for vertical space

\begin{subfigure}[h!]{.5\textwidth} % Adjust the width based on your preference
    \centering
    \includegraphics[width=0.9\textwidth]{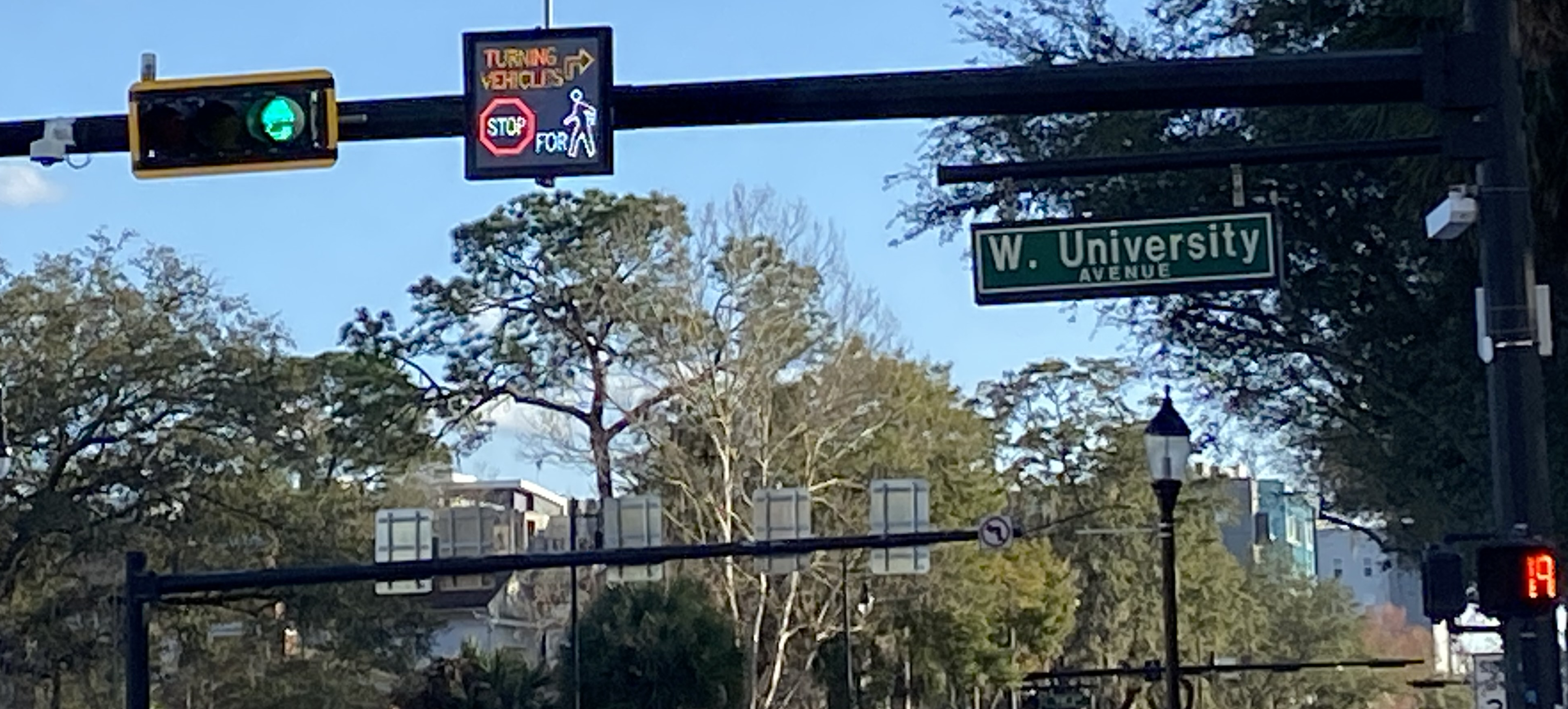}
    \captionsetup{justification=justified, singlelinecheck=false} % Adjust the justification here
    \caption{}
    \label{fig:yield-to-ped-sign}
\end{subfigure}

\caption{University Ave \& 13th St. intersection blank-out signs ``no right turn on red'' and ``turning vehicles stop for pedestrians.''}
\label{fig:intersection-signs}
\end{figure}

This intersection is located on the northeast section of the campus and borders the city's Downtown area to the east, considered to be the central place for pedestrian traffic and the heart of the city.  Two major arterial roads for the city meet at this intersection, carrying an Annual Average Daily Traffic (AADT) of 23,500 - 33,500 along 13th Street and 19,500 - 24,000 along the University Ave \cite{aadt-report}. Notably, this location has been criticized for being dangerous to pedestrians and bicyclists, at times harboring fatal crashes~\cite{henderson}.

As shown in Figure~\ref{fig:pedestrian-path-map}, this intersection is around 3000 feet from the University of Florida's football stadium, which holds around 90,000 people during the university's home games. To prevent extensive traffic congestion during peak times when large numbers of spectators arrive or depart, side streets closer to the stadium are blocked. This operational strategy aims to ensure smoother traffic flow on the main arterial road, facilitate access to the stadium, and enhance the overall transportation experience for both event attendees and regular commuters.%Several measures are taken during such home games at this intersection, including
%\TODO{Emmanuel}{}<please ask Emmanuel what were the measures taken, if any at this location so that it can be highlighted here.>

In Figure~\ref{fig:univ-13-inter}, we present a snapshot of the intersection, showcasing the labeled traffic and pedestrian phases. The fisheye camera is positioned atop the north leg crosswalk, approximately at the midpoint of the crosswalk. As a result, pedestrian phase 4 is prominently captured at this intersection, followed by pedestrian phases 2 and 6. However, due to the camera's fisheye lens and location, the south leg crosswalk is the farthest away from the intersection. Consequently, pedestrians on this crosswalk are often missed, primarily due to their relatively small size and the distortions introduced by the fisheye camera. These challenges in capturing data from the south leg crosswalk highlight the need for careful consideration when interpreting the pedestrian information in this area. The city is in the process of installing an additional camera to cover this crosswalk, which will alleviate these issues in the near future.

Figure~\ref{fig:intersection-signs} shows photographs of two blank-out signs that were implemented within the surveyed intersection in November 2021. The first sign forbids vehicles to turn while the signal is red; the second sign tells vehicles to allow pedestrians to cross before turning. Our analysis surveys traffic interactions that have taken place after the installation of these signs.

To ensure thoroughness in our analysis, we examine traffic patterns throughout the entire day. We divide the time frame into 2-hour intervals, as depicted in Table~\ref{tab:times}. This approach allows us to capture variations in traffic volume and behavior across different segments of the day, enabling a more detailed understanding of traffic dynamics and facilitating data-driven insights.

% is marked as site 5012 according to the Florida Department of Transportation with Description SR 26 (UNIV. AVE.)W.OF NW 13TH ST, Direction 1 E 11000, Direction 2 W 13000, AADT two-way 24000 C, K factor 9.0, D factor 53.3F, t factor 2.0F

% \TODO{ALL}{The timeframes were adjusted to be of equal length (2 hours) because we were recreating the images for conflict volume visualizations. However it is easy to change the timeframe and we can include both total and normalized (conflict per hour) figures if needed. Then we can change the timeframes to be of differing lengths and so they are more realistic to the linguistic definition of "evening", "afternoon" etc. We must discuss this further}

\begin{table}[h] % the [h] means here
\centering
\caption{The ranges of the times of day.}
\bgroup
\def\arraystretch{1.5}
\begin{tabular}{| c | c |}
\hline
{\bf Name} & {\bf Timeframe} \\
\hline
\hline
  Early Morning & 8:00 - 10:00  \\
\hline
  Late Morning & 10:00 - 12:00  \\
\hline
  Early Afternoon & 12:00 - 14:00  \\
\hline
  Late Afternoon & 14:00 - 16:00  \\
\hline
  Evening & 16:00 - 18:00  \\
\hline
  Night & 18:00 - 20:00  \\
\hline
\end{tabular}
\egroup
\label{tab:times}
\end{table}

\subsection{Gamedays Description}
% Which four days, describe them and little bit details about the teams - perhaps have a table or so.
%- Pedestrian path to the stadium should come in here too.
During the collegiate football season, the Florida Gators have
football games on most Saturdays. We analyze four gamedays in 2022 as well as two in 2021; their details are listed in Table~\ref{tab:games}. Additionally, the study includes an analysis of non-gamedays:
\begin{enumerate}
    \item Saturday, September 24, 2022
    \item Saturday, October 1, 2022
    \item Sunday, October 9, 2022
    \item Saturday, October 22, 2022
\end{enumerate}
These days were chosen to represent typical Saturdays and Sundays without football games, allowing for a comparison of traffic patterns and pedestrian activities on regular weekends versus gamedays.

\begin{table*}[h!] % the [h] means here
\caption{The descriptions for the four gamedays in 2022 and two gamedays in 2021. Excitement Index and Home Team Win Probability are rounded to 3 decimal points.}
\centering
\small
\bgroup
\def\arraystretch{1.5}
\begin{tabular}{| c | c | c | c | c | c |}
\hline
\thead{Matchup} & \thead{Date} & \thead{Start\\Time} & \thead{Attendance} & \thead{Excitement\\Index} & \thead{Home Team\\Win Probability} \\
\hline
\hline
  South Florida @ Florida & 2022-09-17 (Sat) & 19:30 & 88,496 & 5.235 & 0.268 \\
\hline
  Eastern Washington @ Florida & 2022-10-02 (Sun) & 12:00 & 72,462 & 1.306 & 0.999 \\
\hline
  Missouri @ Florida & 2022-10-08 (Sat) & 12:00 & 88,471 & 5.102 & 0.930  \\
\hline
  LSU\tablefootnote{Louisiana State University} @ Florida & 2022-10-15 (Sat) & 19:00 & 90,585 & 4.006 & 0.088\\
\hline
\hline
  Samford @ Florida & 2021-11-13 (Sat) & 12:00 & 70,098 & 3.263 & 0.999 \\
\hline
  Florida State @ Florida & 2021-11-27 (Sat) & 12:00 & 88,491 & 6.189 & 0.914 \\
\hline
\end{tabular}
\egroup
\label{tab:games}
\end{table*}

During the data collection for this study, the days selected for analysis were specifically chosen from the months of September and October. These two months have been identified to have the highest number of crashes in Gainesville, according to data from the City of Gainesville~\cite{gnv-traffic-crashes}. By focusing on days within these months, the study aims to identify and analyze any pedestrian-to-vehicle or vehicle-to-vehicle conflicts that may occur during periods of heightened traffic. To increase the amount of gameday data for analysis, two gamedays from the previous year were included in the study. There are no in-house records of gamedays during the months of September and October in the prior year; therefore, two gamedays from November in the prior year were used instead to augment the available data. In any case, November is also of interest as it has the fourth highest number of crashes~\cite{gnv-traffic-crashes}, and the entirety of the three selected months are during the regular football season.

The statistics regarding the gamedays were fetched from the College Football Database (CFBD) and provide key characteristics for further analysis, such as understanding how the pedestrian volume within the intersection fluctuates around the start and end times of the football games. Furthermore, the excitement index, which is a measure of how exciting the game was to watch (calculated by measuring swings in win probability throughout the game~\cite{cfbd}), is also shown in Table~\ref{tab:games}. The attendance, which is the number of spectators in the stadium, is also shown.

By gathering data on the volume of pedestrians, this stage of the paper aims to determine the extent of the pedestrian volume increase and the pedestrians' movements due to a nearby football game.

%\pagebreak

\subsection{Heatmaps}

A way to visualize the change in pedestrian and vehicle volume over the course of one day is to construct heatmaps of phase over time, as shown in Figures~\ref{fig:ped-vol-heatmaps} through \textbf{\ref{fig:non-gameday-veh-vol-heatmaps}}. The heatmap grids vary in color intensity to show the variation in volume. Furthermore, if the day showcased in the heatmap is a gameday, then there are football icons to signify the start and end of a game. These heatmaps are augmented with the CFBD API---specifically their Python module, \textit{cfbd}, that returns the statistics of a football game on a particular date~\cite{cfbd}. This API provides the times that a football game begins, which is of interest because the football game spectators will affect the intersection pedestrian volumes. To identify the correlation between football match start times and pedestrian volume, the days listed within the CSV file are used to query the CFBD database. Also, the end time is estimated using the average time of a Football Bowl Subdivision (the division of the Florida Gators) game, which is around 3 hours and 22 minutes~\cite{parks, smits}.

% \TODO{JPF}{take into account how prominent the team is, the cfbd has a statistic}

\subsection{Volume Analysis}
%insert heatmap and describe observations
In this study, we commence our analysis by presenting an evaluation of pedestrian volume at the selected intersection. We focus on various time frames, including gamedays and non-gamedays, to understand pedestrian volumes and behavior during different scenarios. The pedestrian volume analysis includes observations on peak hours, fluctuations in pedestrian numbers, and any noticeable patterns related to special events or weather conditions.

Subsequently, we shift our focus to vehicle volume analysis at the same intersection. We examine traffic patterns during both gamedays and non-gamedays to gain insights into vehicle volumes.

\subsubsection{Pedestrian Volume Analysis}

\begin{figure*}[ht!]
\centering

\begin{minipage}[b]{0.45\textwidth}
\centering
\includegraphics[height=4.9cm]{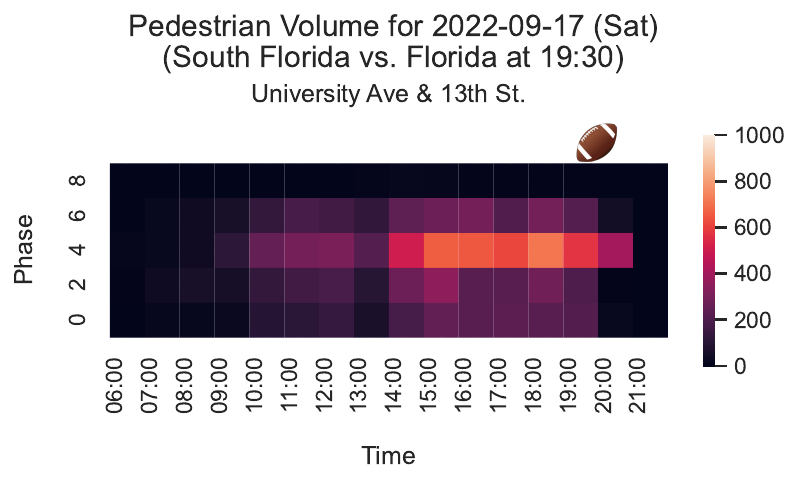}
\end{minipage}
\hfill
\begin{minipage}[b]{0.45\textwidth}
\centering
\includegraphics[height=4.9cm]{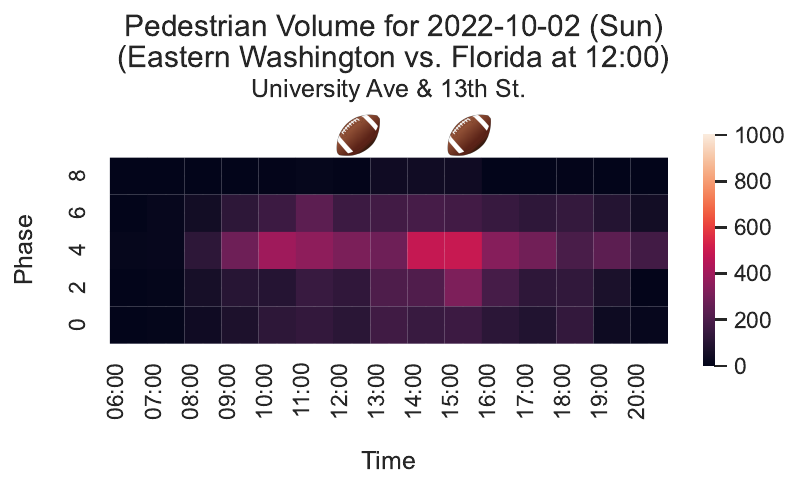}
\end{minipage}

\vspace{0.2cm}

\begin{minipage}[b]{0.45\textwidth}
\centering
\includegraphics[height=4.9cm]{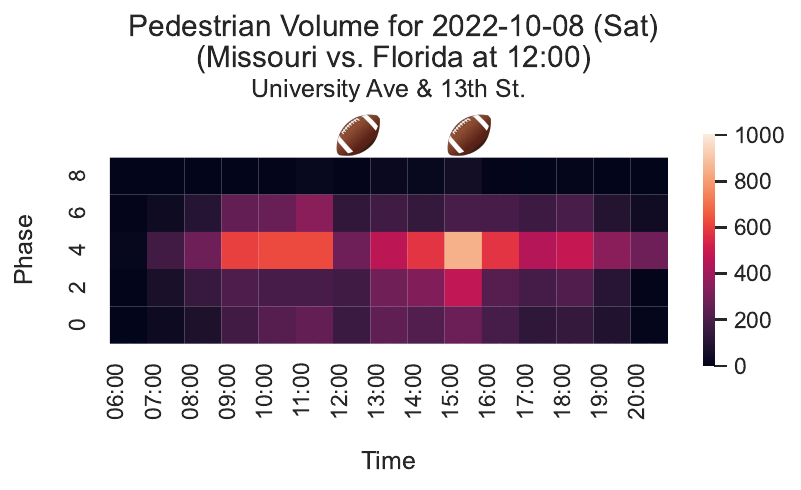}
\end{minipage}
\hfill
\begin{minipage}[b]{0.45\textwidth}
\centering
\includegraphics[height=4.9cm]{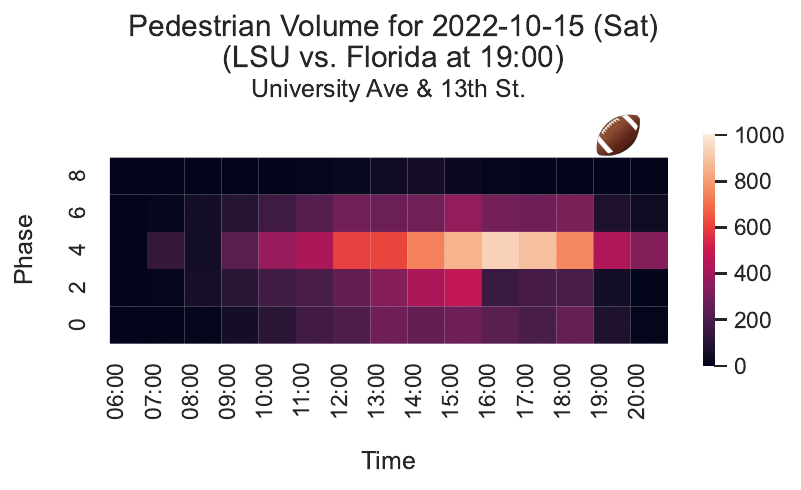}
\end{minipage}

\caption{Heatmaps for four gamedays that show the density of pedestrian volume as well as the football game start time. The leftmost position of the football icons shows the start or end time of the game. The games on 2022-09-17 and 2022-10-15 ended later than 9 PM; hence, the game-end football icons are missing.}\label{fig:ped-vol-heatmaps}
\end{figure*}

Figure~\ref{fig:ped-vol-heatmaps} shows the pedestrian volume for four gamedays in 2022. The figure contains football icons to signify the start and end times of the game along the x-axis, which represents time. If the football game's start and end were recorded within the data's timeframe (we have video data until 9 PM for most days), then both corresponding icons will show in the figure. However, if the end of the game is too late into the night, the second football icon to represent the end is not included, as the camera cannot record well in the night due to obscurity.

\begin{figure*}[ht!]
\centering

\begin{minipage}[b]{0.45\textwidth}
\centering
\includegraphics[height=4.9cm]{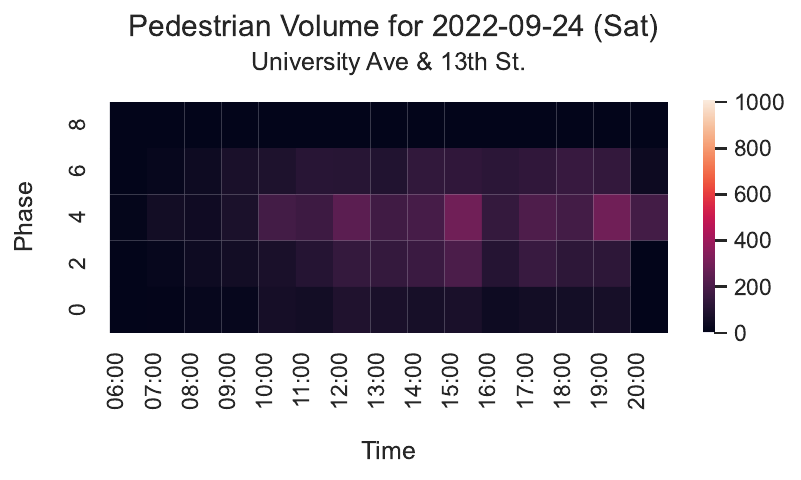}
\end{minipage}
\hfill
\begin{minipage}[b]{0.45\textwidth}
\centering
\includegraphics[height=4.9cm]{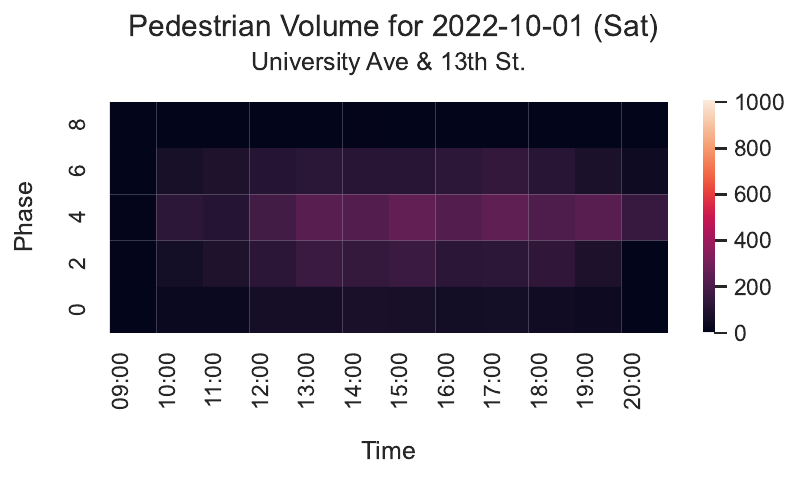}
\end{minipage}

\vspace{0.2cm}

\begin{minipage}[b]{0.45\textwidth}
\centering
\includegraphics[height=4.9cm]{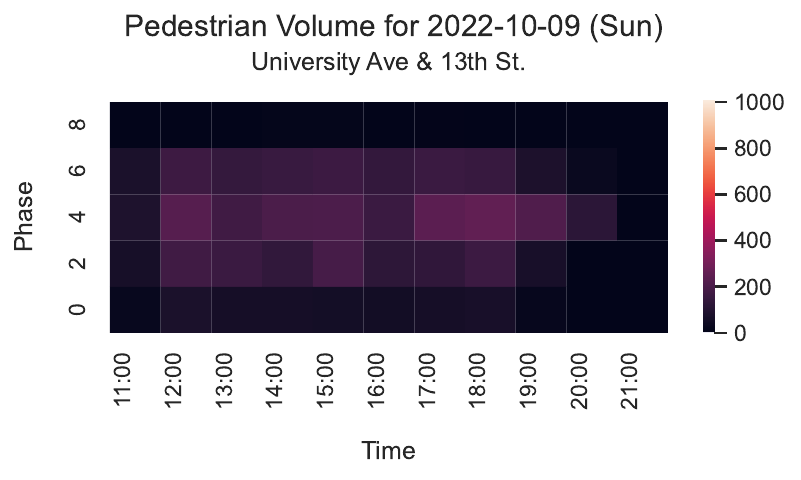}
\end{minipage}
\hfill
\begin{minipage}[b]{0.45\textwidth}
\centering
\includegraphics[height=4.9cm]{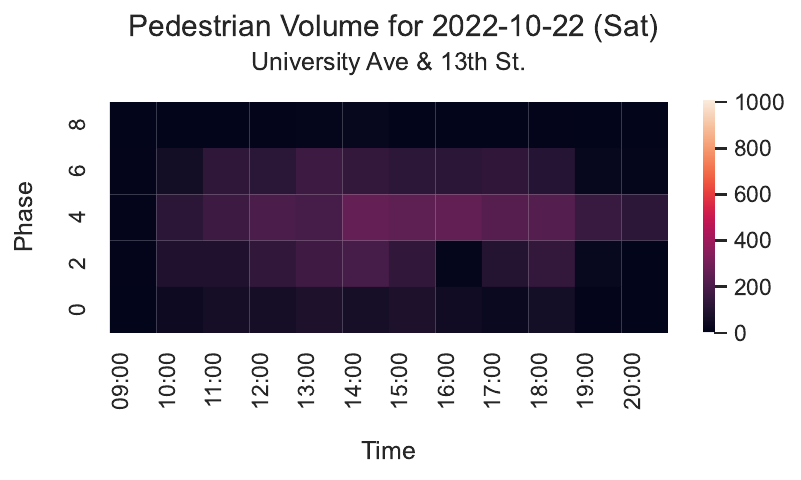}
\end{minipage}

\caption{Heatmaps for four non-gamedays that show the density of pedestrian volume.}\label{fig:non-game-days-ped-vol-heatmaps}
\end{figure*}

Similarly, Figure~\ref{fig:non-game-days-ped-vol-heatmaps} shows the pedestrian volume for four non-gamedays in 2022 for the same intersection. Both analyses include three Saturdays and one Sunday. There is a pronounced difference in pedestrian volume in Figure~\ref{fig:ped-vol-heatmaps} as compared to Figure~\ref{fig:non-game-days-ped-vol-heatmaps}; whereas non-gamedays maximum volume is around 400-500 per hour, the maximum volume for gamedays is near 1,000 pedestrians per hour.

The pedestrian volumes show correlations with weather events. For instance, on the second gameday on October 2, 2022, which occurred on a Sunday, fewer pedestrians were present in the intersection. This decrease in pedestrian activity can be attributed to Hurricane Ian passing through Florida between September 28 and 30, 2022. As a result of the weather conditions, the game that was originally scheduled for Saturday, October 1, was postponed to Sunday, October 2, 2022. The impact of the hurricane led to a dampened turnout on the rescheduled gameday, resulting in reduced pedestrian presence at the intersection.

In Figure~\ref{fig:ped-vol-heatmaps}, the grids leading up to a game possess an elevated pedestrian volume level, and the grid immediately after (during the game) shows a drop in volume. The most voluminous phase is Phase 4. Furthermore, three hours after the game's start time, there is a slight uptick in Phase 2 volume. This uptick could be due to pedestrians returning to a nearby hotel or a parking garage after the football game ended; Phase 2 is from south to north, so they are retracing their steps and crossing Phase 4 again by first crossing Phase 2. Phase 4 is the most voluminous because pedestrians travel from east to west towards the stadium and then again from west to east. On all days, Phase 4 is the most-used pedestrian phase, and Phase 6 is the least-used phase. 

% football image https://publicdomainvectors.org/en/free-clipart/American-football-ball-vector-image/14669.html

%Instead of simply looking at the fluctuations in volume, some characteristics of the football game itself may have correlations with the volume.

\subsubsection{Vehicle Volume Analysis}
%insert heatmap and describe observations

\begin{figure*}[ht!]
\centering

\begin{minipage}[b]{0.45\textwidth}
\centering
\includegraphics[height=5cm]{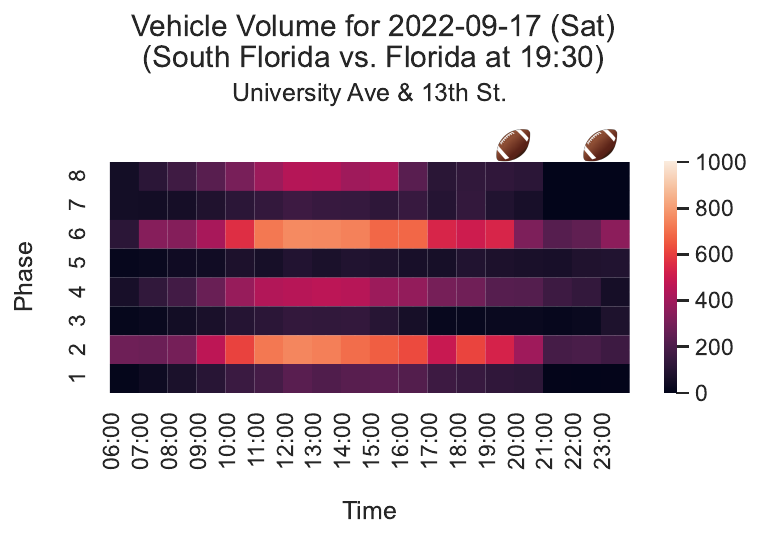}
\end{minipage}
\hfill
\begin{minipage}[b]{0.45\textwidth}
\centering
\includegraphics[height=5cm]{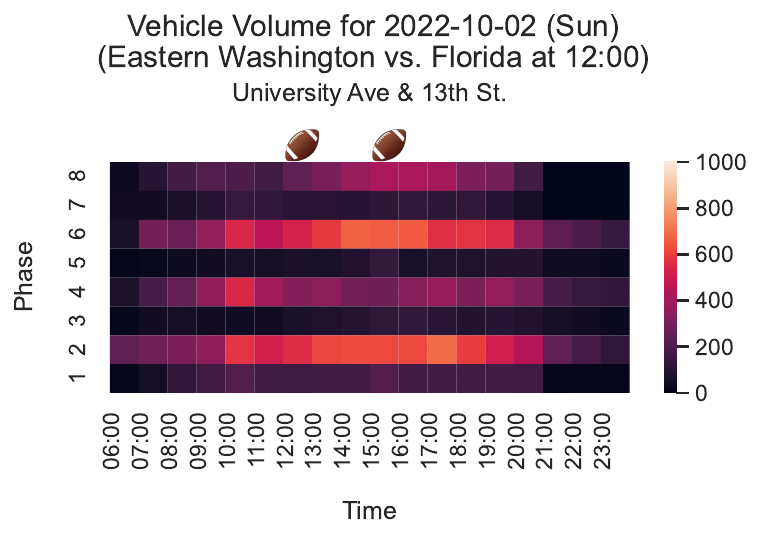}
\end{minipage}

\vspace{0.1cm}

\begin{minipage}[b]{0.45\textwidth}
\centering
\includegraphics[height=5cm]{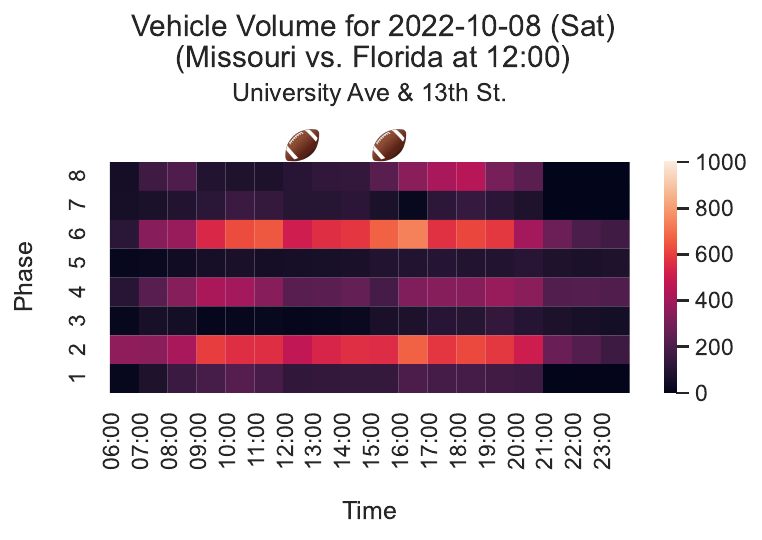}
\end{minipage}
\hfill
\begin{minipage}[b]{0.45\textwidth}
\centering
\includegraphics[height=5cm]{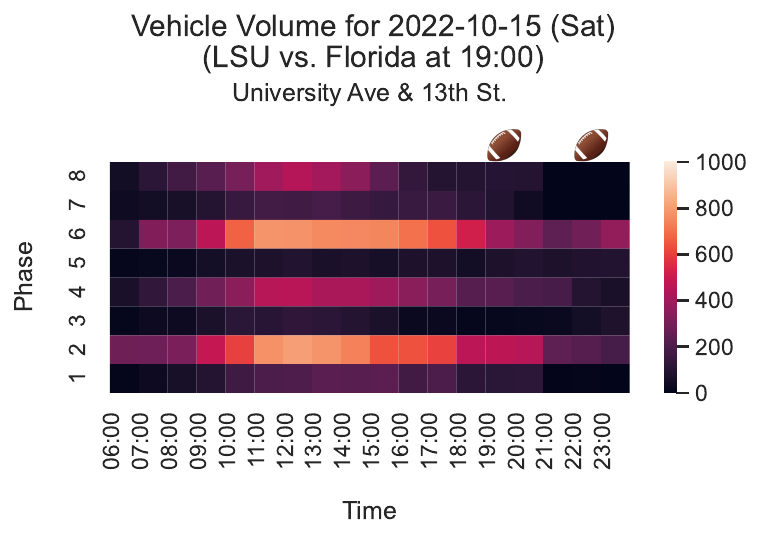}
\end{minipage}

\caption{Heatmaps for four gamedays that show the density of vehicle volume as well as the football game start time. The leftmost position of the football icons shows the start or end time of the game.}\label{fig:veh-vol-heatmaps}
\end{figure*}

\begin{figure*}[ht!]
\centering

\begin{minipage}[b]{0.45\textwidth}
\centering
\includegraphics[height=4cm]{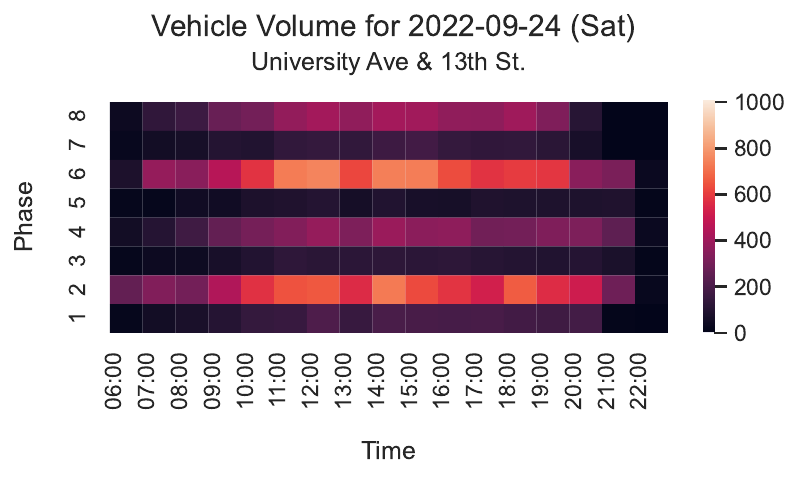}
\end{minipage}
\hfill
\begin{minipage}[b]{0.45\textwidth}
\centering
\includegraphics[height=4cm]{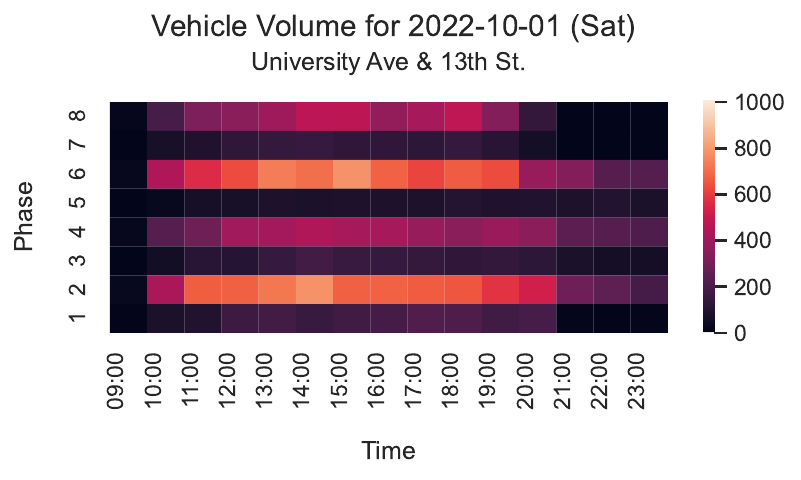}
\end{minipage}

\vspace{0.1cm}

\begin{minipage}[b]{0.45\textwidth}
\centering
\includegraphics[height=4cm]{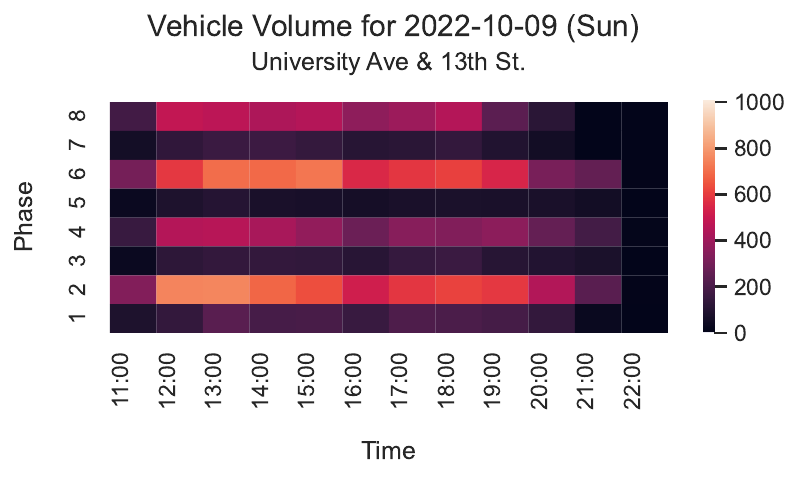}
\end{minipage}
\hfill
\begin{minipage}[b]{0.45\textwidth}
\centering
\includegraphics[height=4cm]{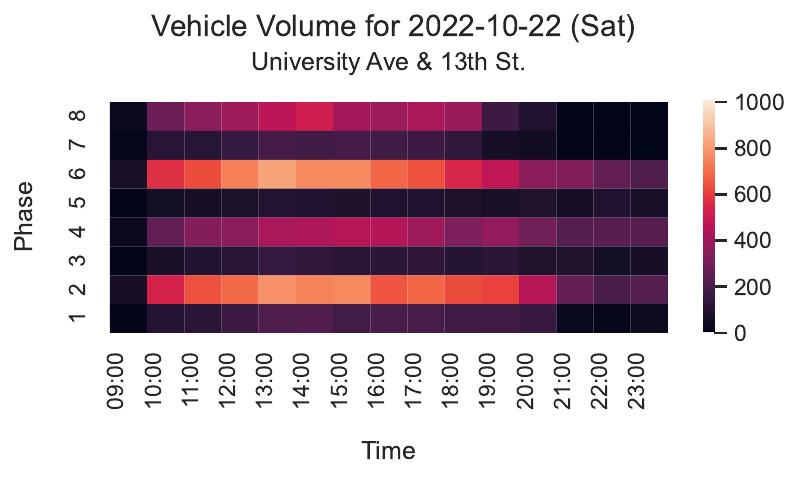}
\end{minipage}

\caption{Heatmaps for four non-gamedays that show the density of vehicle volume.}\label{fig:non-gameday-veh-vol-heatmaps}
\end{figure*}

The heatmaps in Figures~\ref{fig:veh-vol-heatmaps}~and~\ref{fig:non-gameday-veh-vol-heatmaps} contain eight different phases to portray all the possible movements and trajectories that vehicles can take within the intersection, as referenced in Figure~\ref{fig:phases}.

The gamedays, especially featuring more popular matchups such as the visiting South Florida and LSU teams, see more cells that reach the highest-seen vehicle count of $\sim$900 vehicles per hour. Non-gamedays do not often reach this maximum; if they do, it is not continuous over several hours.

Furthermore, games that end late in the day contain an uptick in vehicle volume immediately after the game's end, likely due to fans leaving and returning to their cars. Also, when gamedays have a starting time early in the day, there is an immediate drop in vehicle volume that is not seen in non-gamedays.

%\pagebreak

\begin{figure*}[htp]
\centering
\begin{subfigure}[b]{0.40\textwidth}
  \centering
  \includegraphics[height=7cm]{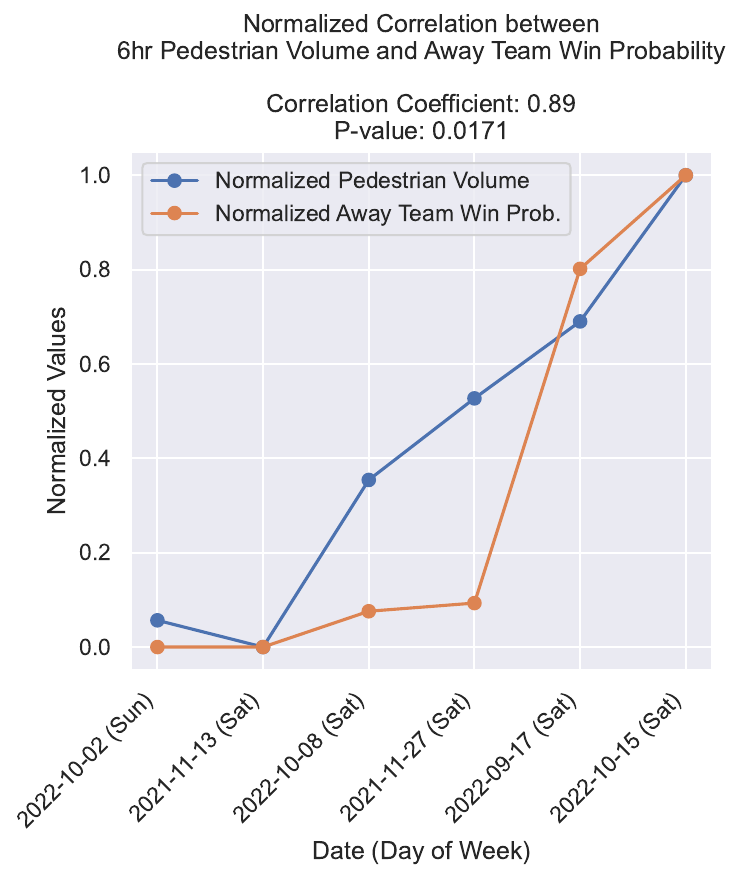}
  \captionsetup{justification=centering}
  \caption{}
  \label{fig:ped-vol-win-prob}
\end{subfigure}
% \hfill
\begin{subfigure}[b]{0.40\textwidth}
  \centering
  \includegraphics[height=7cm]{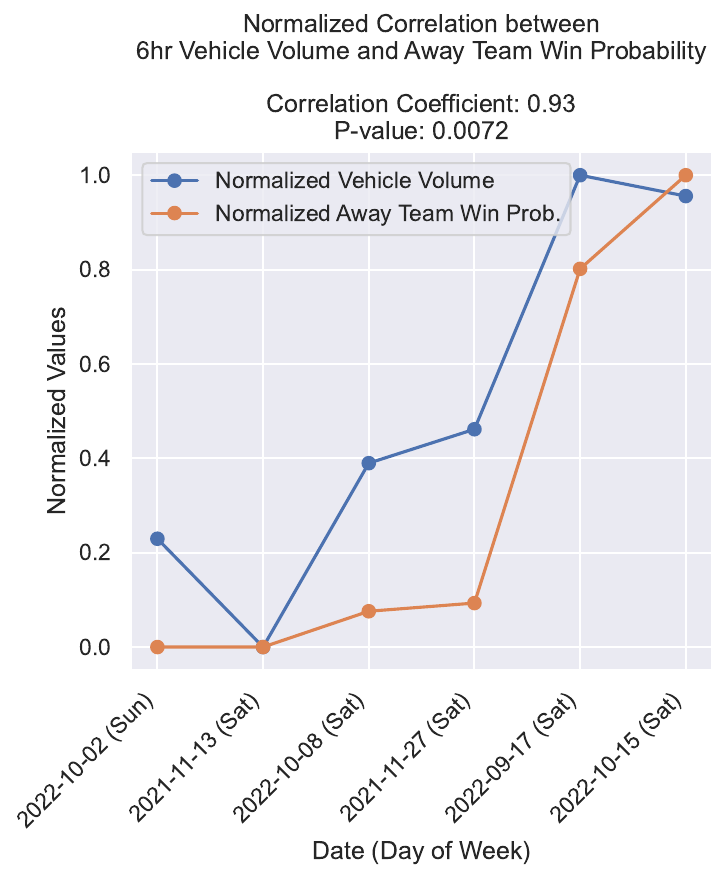}
  \captionsetup{justification=centering}
  \caption{}
  \label{fig:veh-vol-win-prob}
\end{subfigure}
\caption{Correlation between volume and away team win probability for both pedestrian and vehicle traffic within the intersection for 6 hours leading up to the game.}
\label{fig:correlation-vol-win-prob}
\end{figure*}
\begin{figure*}[h!]
\centering
\begin{subfigure}[h!]{.45\textwidth}
    \centering
    \includegraphics[width=0.85\columnwidth]{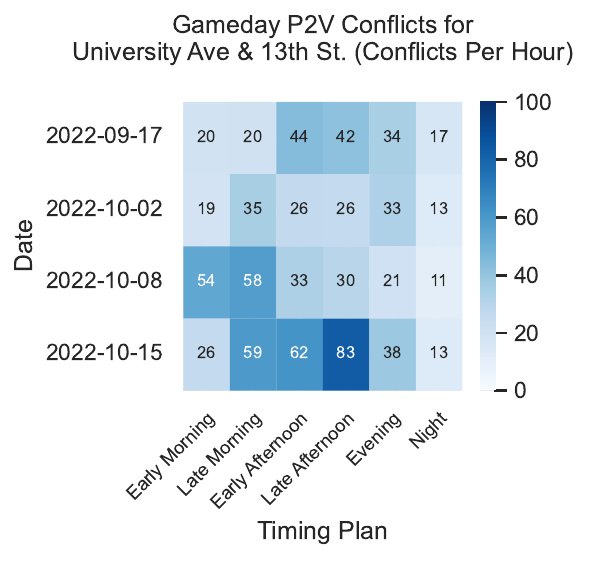}
    \captionsetup{justification=centering}
    \caption{}
    \label{fig:p2v-normalized-heatmap}
\end{subfigure}
\hspace{1cm} % Adjust the value as needed for the desired horizontal space
\begin{subfigure}[h!]{.45\textwidth}
    \centering
    \includegraphics[width=0.85\columnwidth]{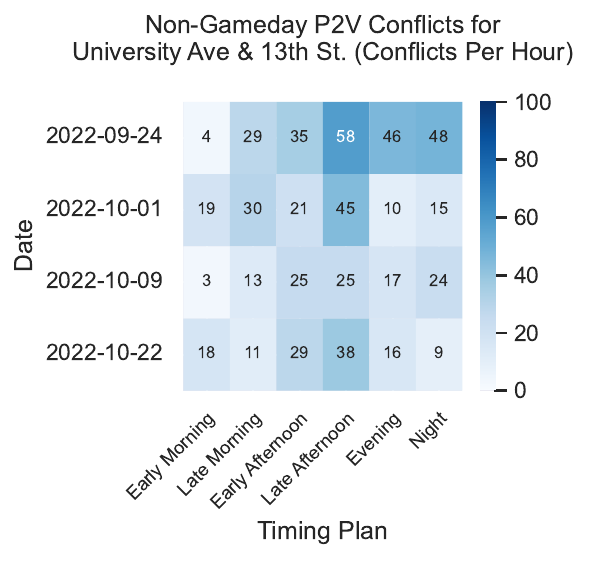}
    \captionsetup{justification=centering}
    \caption{}
    \label{fig:p2v-non-gameday-normalized-heatmap}
\end{subfigure}
\caption{Heatmaps that show the total P2V conflicts for both gamedays and non-gamedays.}
\label{fig:p2v-normalized-conflicts}
\end{figure*}
\subsection{Win Probability Correlation}

To find a relationship between the pedestrian volume and the away team win probability, the pedestrian volume count was determined by adding the total pedestrians identified within the intersection for six hours up to the game start. Six hours instead of some other number is chosen because we anticipate that football fans may frequent the area up to six hours earlier for pre-game activities, such as dining or entertainment. The away team win probability is queried from the CFBD API. If desired, the home team win probability can be chosen instead, with the only effect having a reverse correlation (but the coefficient absolute value remains the same). 

Firstly, min-max normalization is applied to both datasets (the pedestrian volume and away team win probability) using the following, where $x$ is the data and $n$ is the total number of data points (as shown in Equation~\ref{Eqn:normalization}).

\begin{align}
\label{Eqn:normalization}
   \text{Normalized data} = \biggl[ 
   & \frac{x_1 - \min(\text{data})}{\max(\text{data}) - \min(\text{data})}, \nonumber \\
   & \frac{x_2 - \min(\text{data})}{\max(\text{data}) - \min(\text{data})}, \nonumber \\
   & \quad \vdots \quad \nonumber \\
   & \frac{x_n - \min(\text{data})}{\max(\text{data}) - \min(\text{data})} \biggr]
\end{align}

\vspace{1em}

Normalizing the data causes its range to become $[0,1]$.

Then, the correlation coefficient is calculated and represented as $r$, the mean pedestrian count is $\bar{X}$, and the mean away team win probability is $\bar{Y}$, as indicated in Equation~\ref{Eqn:pearson}. The p-value is evaluated as shown in Equation~\ref{Eqn:p-value}.

\begin{figure}[ht]
\begin{minipage}{0.45\textwidth}
\begin{equation}
r = \frac{\sum_{i=1}^n (x_i - \bar{X})(y_i - \bar{Y})}{\sqrt{\sum_{i=1}^n (x_i - \bar{X})^2 \sum_{i=1}^n (y_i - \bar{Y})^2}}
\label{Eqn:pearson}
\end{equation}
\end{minipage}\hfill
\begin{minipage}{0.45\textwidth}
\begin{equation}
\text{P-value} = P(|r| \geq |r_{\text{observed}}|)
\label{Eqn:p-value}
\end{equation}
\end{minipage}
\end{figure}

After these calculations, the program creates a line plot. The results are shown in Figure~\ref{fig:correlation-vol-win-prob}. Specifically, a correlation coefficient of 0.89 and p-value of 0.0171 is found between 6-hour pedestrian volume and away team win probability. Similarly for vehicles, a correlation coefficient of 0.93 and p-value of 0.0072 is found between 6-hour vehicle volume and away team win probability. In both cases, the high coefficients and low p-values indicate a significant correlation that is unlikely due to chance. Thus, Figure~\ref{fig:correlation-vol-win-prob} suggests that researchers and traffic engineers can use the probability of the home team winning to make significant changes to the signal timings and its corresponding parameters. Cross-referencing this correlation with Table~\ref{tab:games} shows that the games with a higher attendance tend to correspond with games where a Florida Gators win is less certain. Therefore, using the win probability can determine the pedestrian and vehicle volume within the intersection preemptively. This is likely because fans enjoy watching rivalries or matchups that involve a more difficult game.

\subsection{Conflict Analysis}
In this section, we delve into an analysis of both P2V and V2V conflicts at the selected intersection. For both analyses, we initially provide an aggregated count of conflicts observed during the data collection period. This quantitative assessment helps establish a clear understanding of the frequency and magnitude of conflicts that occur at the intersection. Subsequently, we conduct an in-depth analysis to explore any potential correlations between these conflicts and the occurrence of football games.

\begin{figure}[htp]
\centering
\includegraphics[height=6cm]{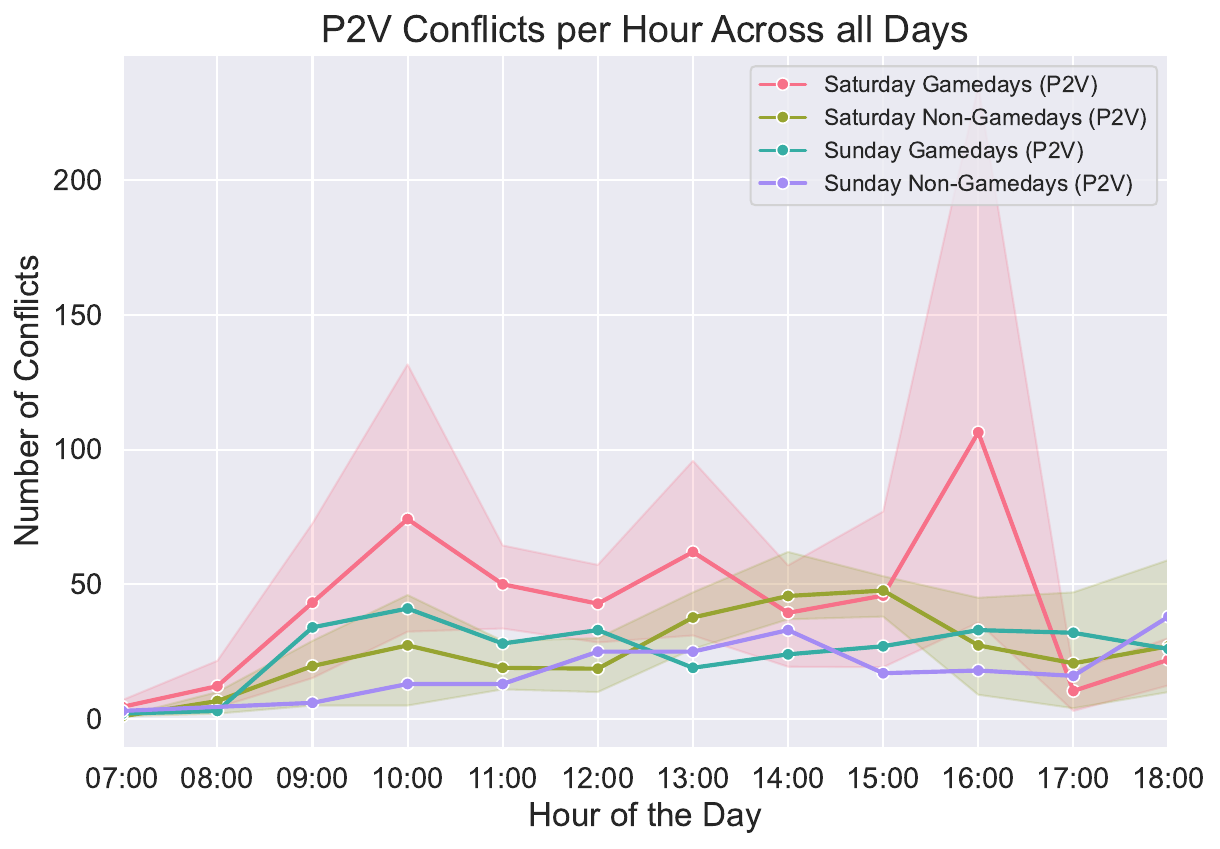}
\caption{Aggregated P2V conflicts from 07:00 to 18:00.}
\label{fig:aggregated-p2v}
\end{figure}

\begin{figure*}[htp]
\centering
\begin{subfigure}[t]{.45\textwidth}
    \centering
    \includegraphics[width=0.78\columnwidth]{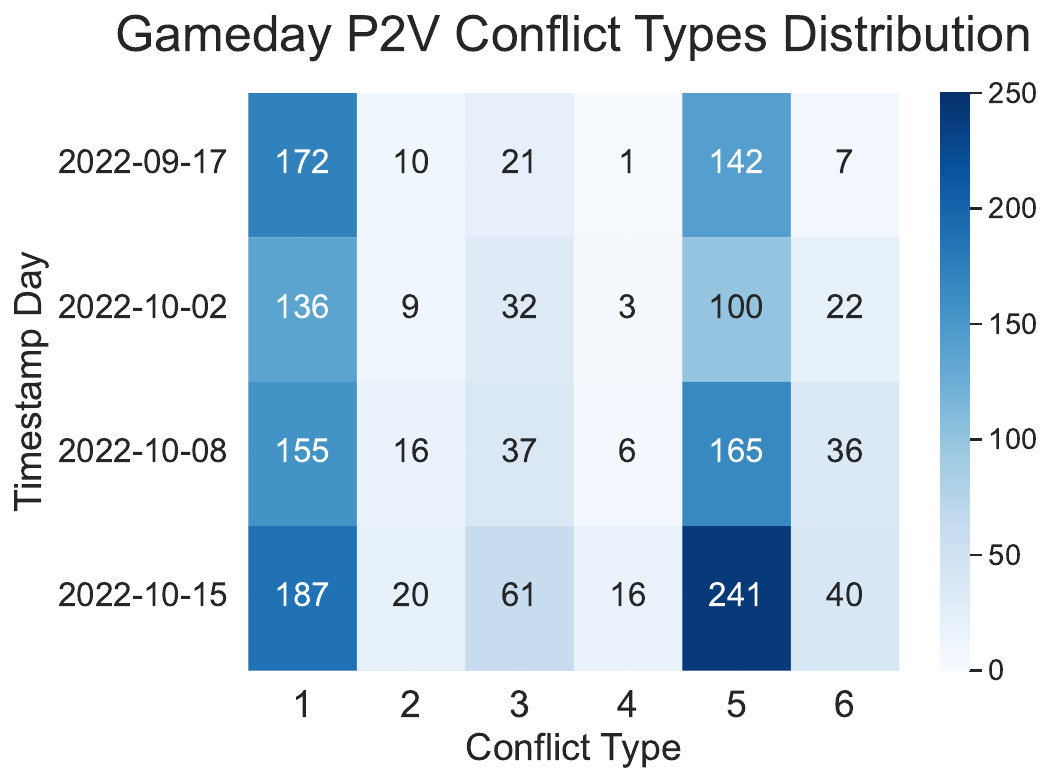}
    \captionsetup{justification=centering}
    \caption{}
    \label{fig:gameday-p2v-conflicts}
\end{subfigure}
\hspace{1cm} % Adjust the value as needed for the desired horizontal space
\begin{subfigure}[t]{.45\textwidth}
    \centering
    \includegraphics[width=0.8\columnwidth]{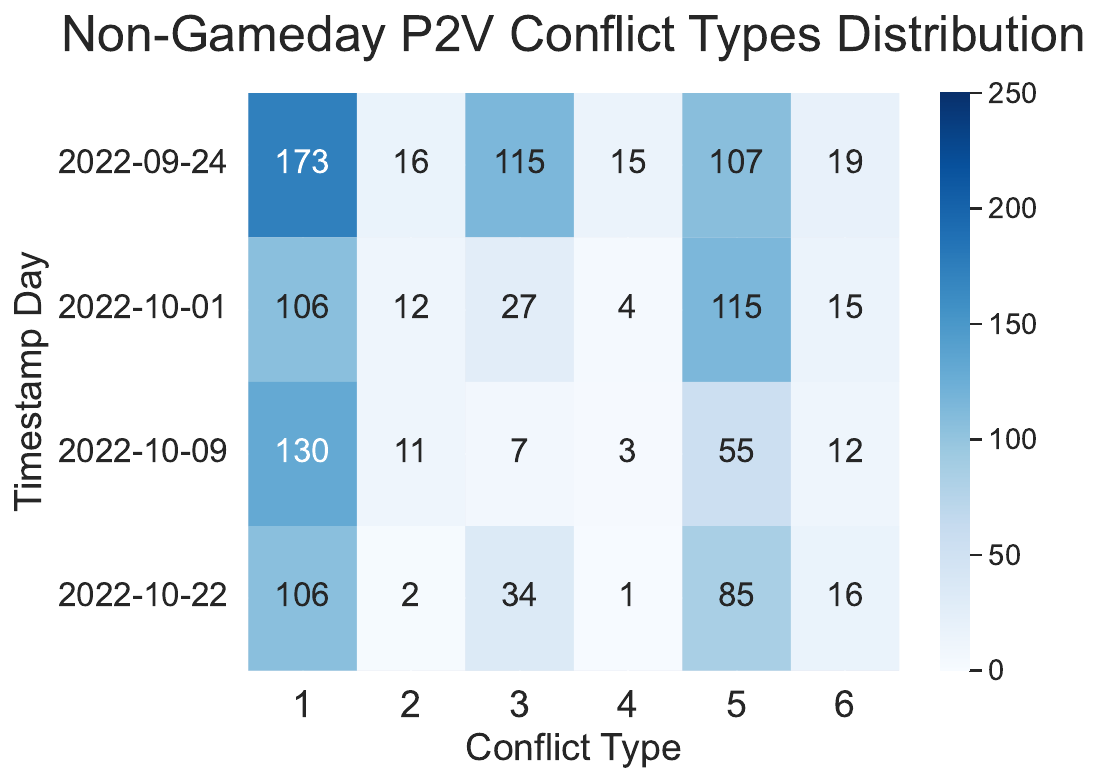}
    \captionsetup{justification=centering}
    \caption{}
    \label{fig:p2v-conflict-types-nongameday}
\end{subfigure}
\caption{Conflict type comparison between gamedays and non-gamedays for pedestrian-to-vehicle conflicts.}
\label{fig:conflict-type-comparison}
\end{figure*}
\begin{figure*}[h!]
\centering

% First Row
\begin{minipage}[b]{0.45\textwidth}
  \centering
  \includegraphics[height=7cm]{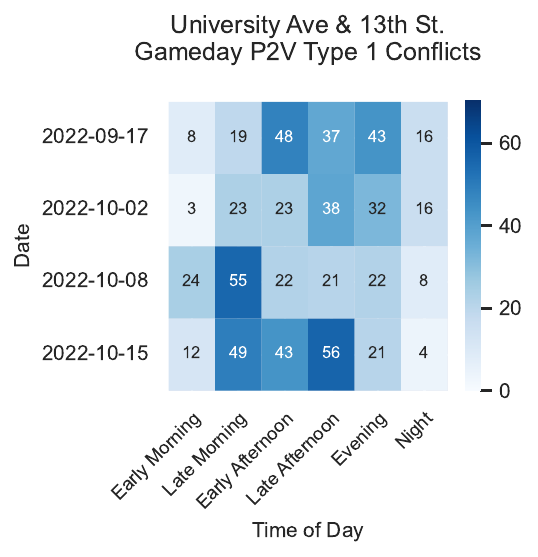}
\end{minipage}
\hspace{0.02\textwidth} % Adjust this value as needed to shift the figure
\begin{minipage}[b]{0.45\textwidth}
  \centering
  \includegraphics[height=7cm]{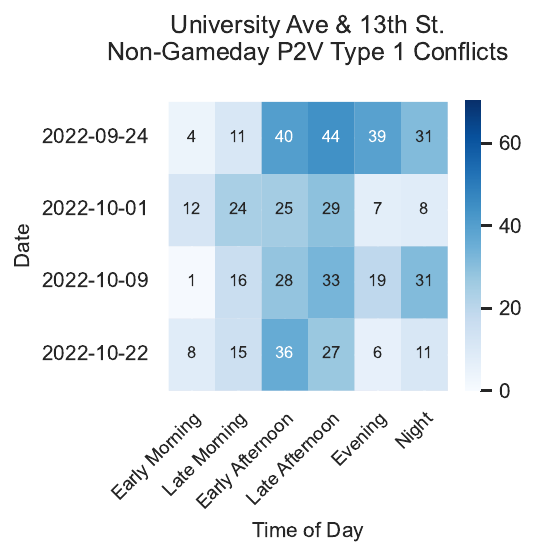}
\end{minipage}

% Second Row
\begin{minipage}[b]{0.45\textwidth}
  \centering
  \includegraphics[height=7cm]{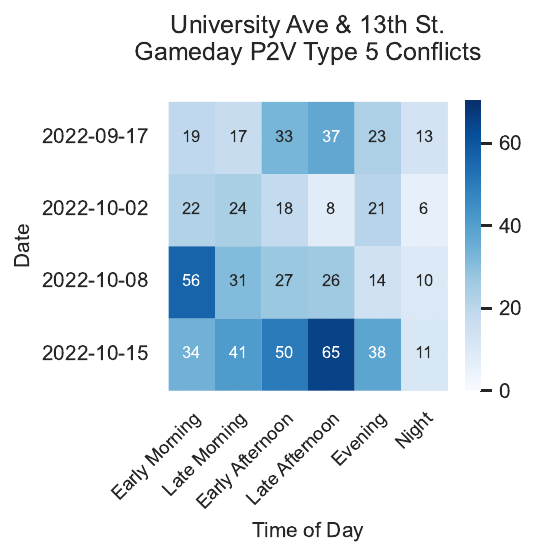}
\end{minipage}
\hspace{0.02\textwidth} % Adjust this value as needed to shift the figure
\begin{minipage}[b]{0.45\textwidth}
  \centering
  \includegraphics[height=7cm]{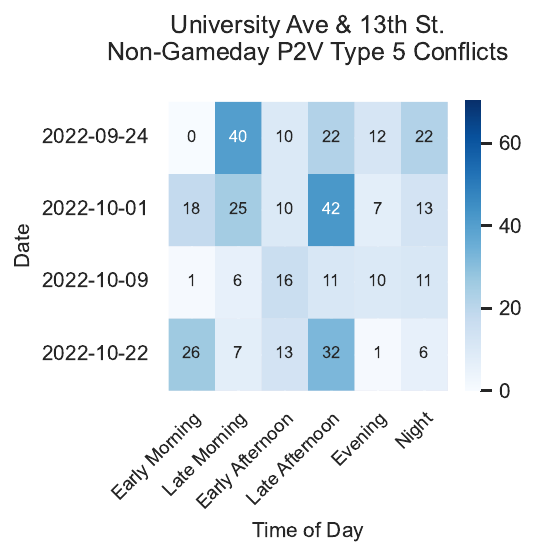}
\end{minipage}

\caption{Heatmaps for two P2V types for four days (two gamedays and two non-gamedays). The y-axis shows each day's date, and the x-axis shows each time-of-day. The numerical conflict classification uses a standard previously described in Figure~\ref{fig:conflict-types}. Type 5 is a more severe conflict involving a through vehicle perpendicularly crossing a pedestrian's path.}\label{fig:indiv-type-heatmaps}
\end{figure*}
\begin{figure*}[ht!]
\centering

\begin{minipage}[b]{0.45\textwidth}
  \centering
  \includegraphics[width=0.8\textwidth]{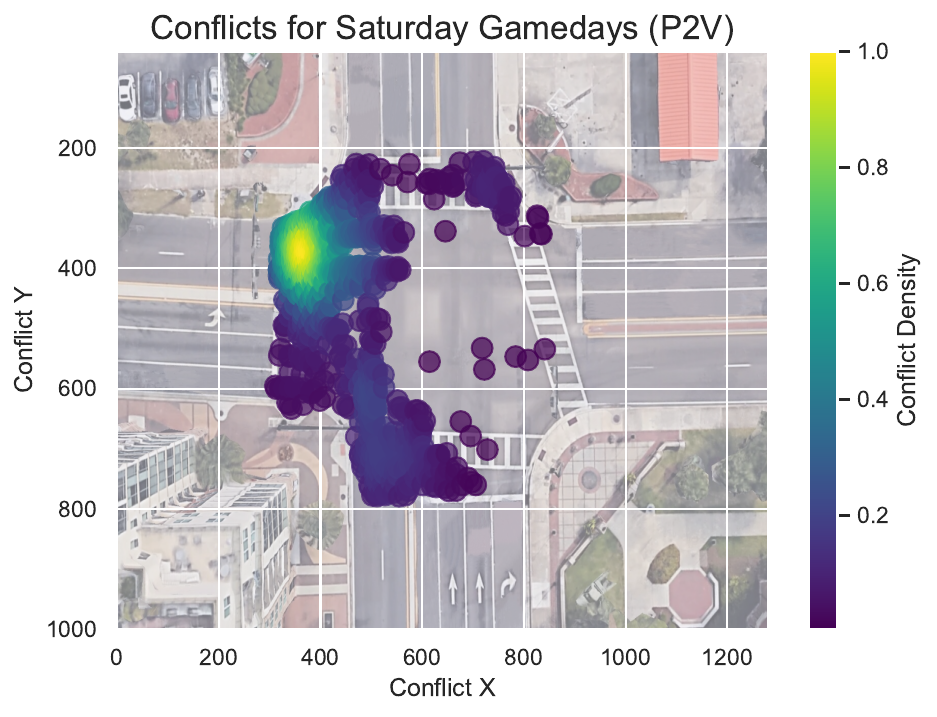}
\end{minipage}
\hfill
\begin{minipage}[b]{0.45\textwidth}
  \centering
  \includegraphics[width=0.8\textwidth]{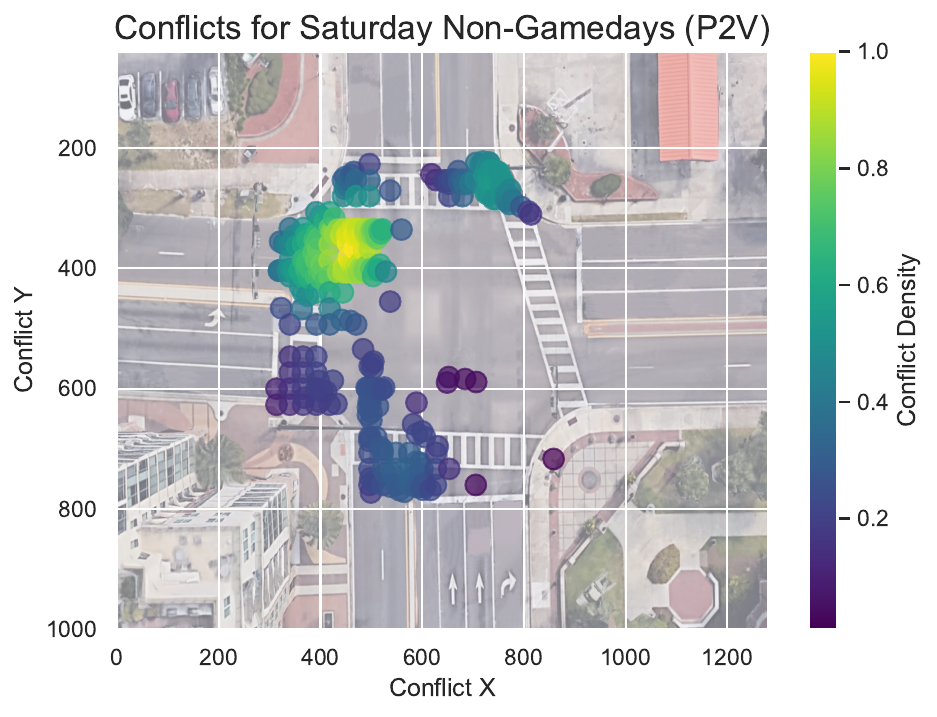}
\end{minipage}

\vspace{0.1cm}

\begin{minipage}[b]{0.45\textwidth}
  \centering
  \includegraphics[width=0.8\textwidth]{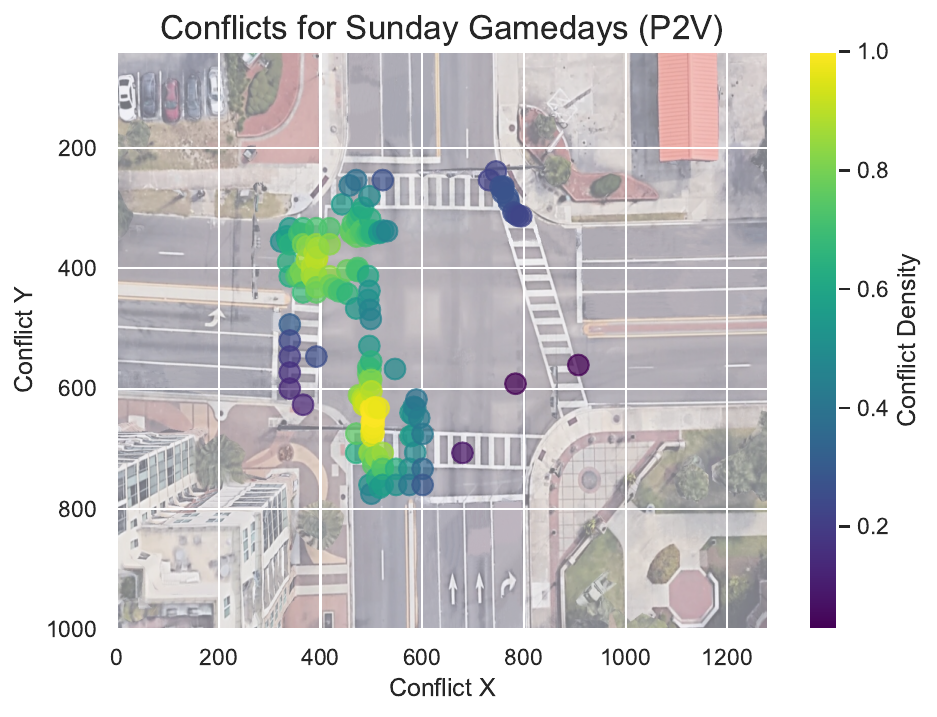}
\end{minipage}
\hfill
\begin{minipage}[b]{0.45\textwidth}
  \centering
  \includegraphics[width=0.8\textwidth]{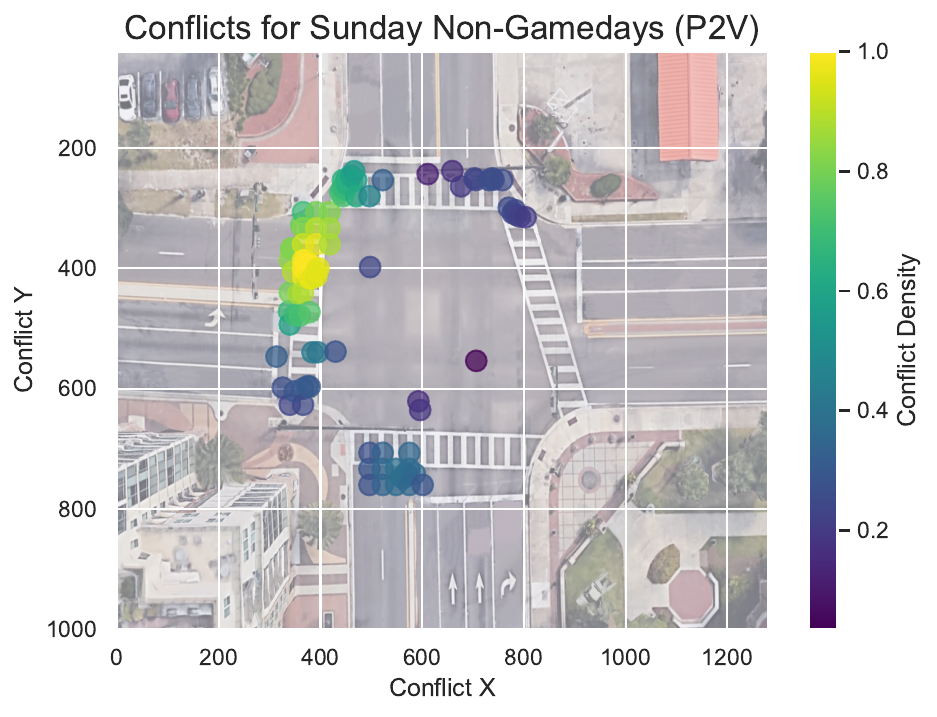}
\end{minipage}

\caption{Spatial distribution scatterplot of P2V conflicts demonstrating the conflict density and clustering within the intersection.}\label{fig:intersec-heatmap-p2v}
\end{figure*}
\begin{figure*}[ht!]
\centering

\begin{minipage}[b]{0.45\textwidth}
  \centering
  \includegraphics[width=0.8\textwidth]{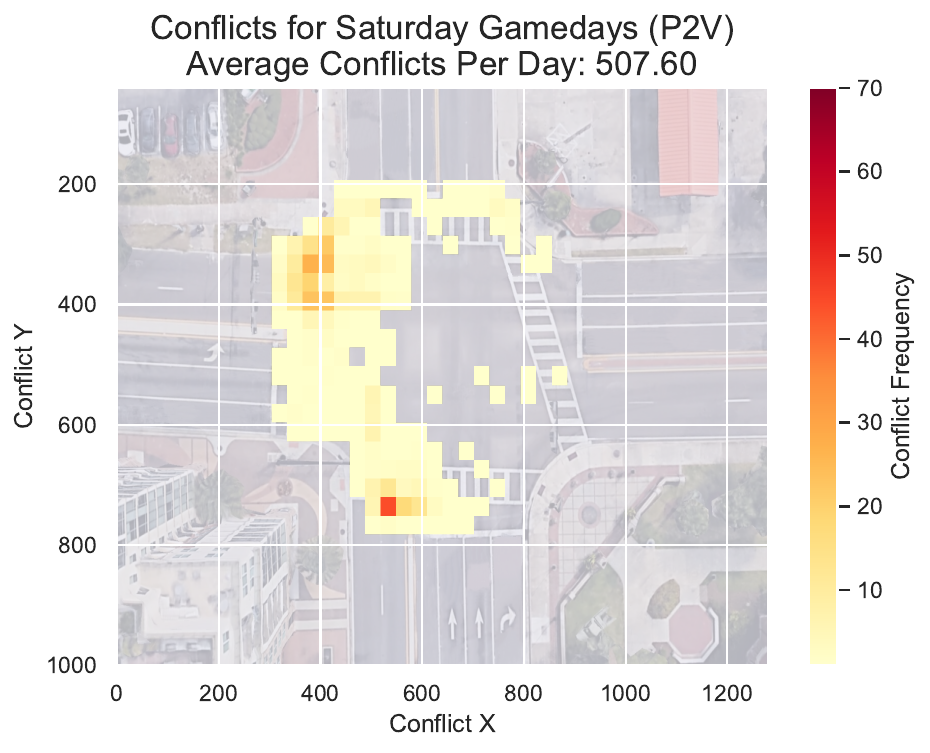}
\end{minipage}
\hfill
\begin{minipage}[b]{0.45\textwidth}
  \centering
  \includegraphics[width=0.8\textwidth]{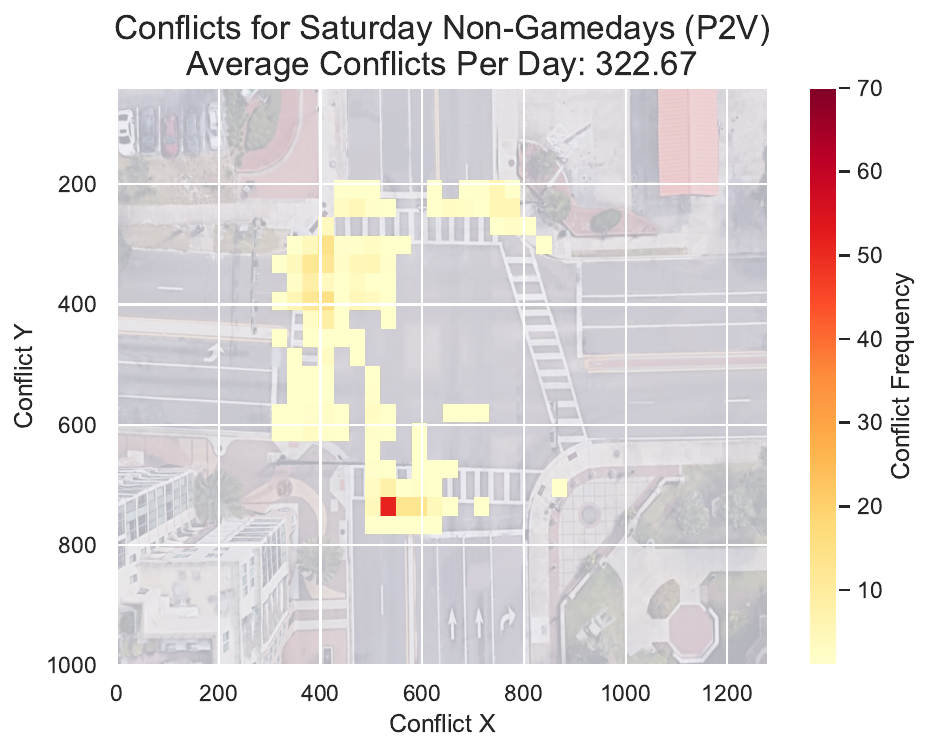}
\end{minipage}

\vspace{0.1cm}

\begin{minipage}[b]{0.45\textwidth}
  \centering
  \includegraphics[width=0.8\textwidth]{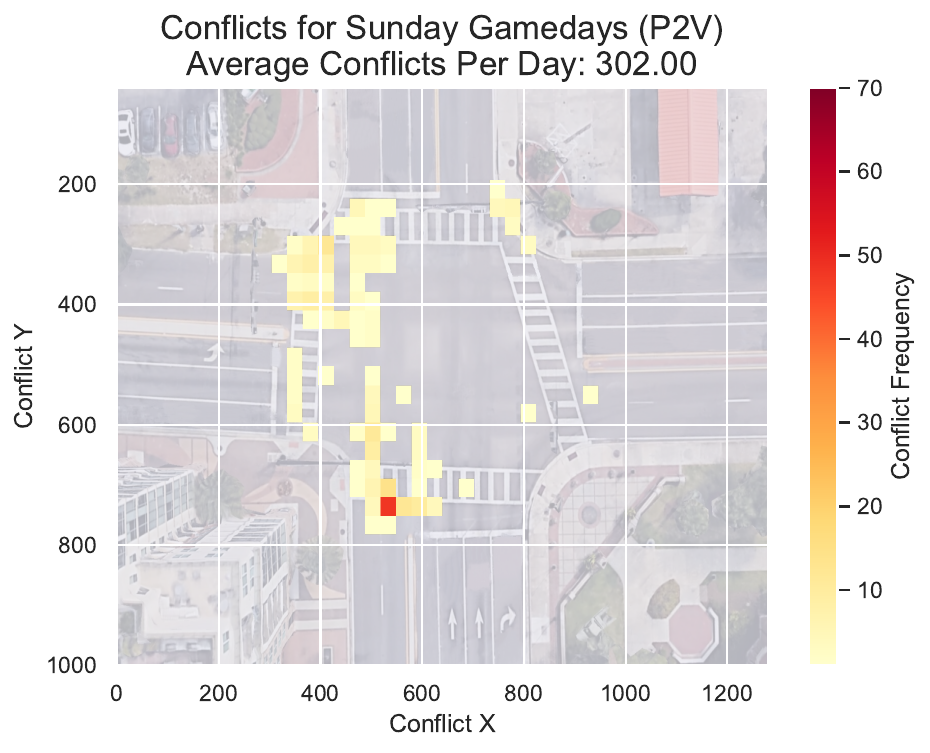}
\end{minipage}
\hfill
\begin{minipage}[b]{0.45\textwidth}
  \centering
  \includegraphics[width=0.8\textwidth]{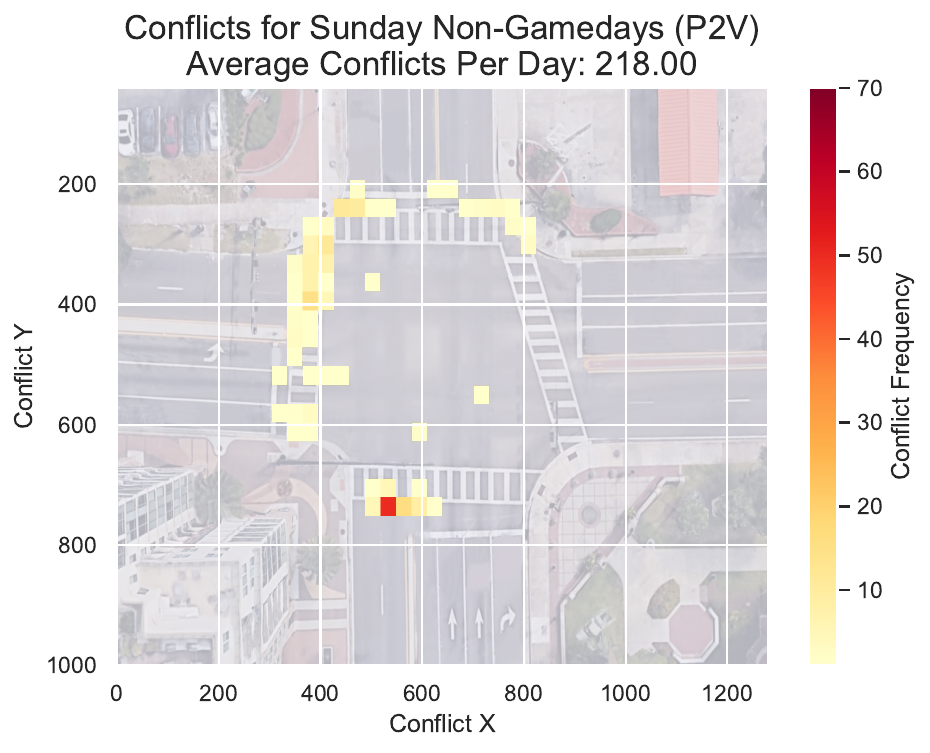}
\end{minipage}

\caption{Spatial distribution histogram of P2V conflicts demonstrating the number of conflicts and their frequency within the intersection.}\label{fig:histogram-heatmap-p2v}
\end{figure*}
\begin{figure*}[htp]
\centering
\begin{subfigure}[t]{.35\textwidth}
    \centering
    \includegraphics[width=0.75\columnwidth]{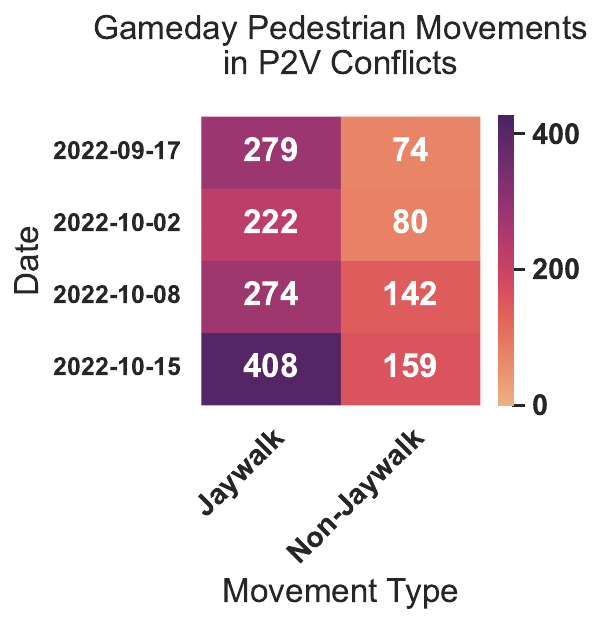}
    \captionsetup{justification=centering}
    \caption{}
    \label{fig:p2v-conflict-ped-movements}
\end{subfigure}
\hspace{1cm} % Adjust the value as needed for the desired horizontal space
\begin{subfigure}[t]{.35\textwidth}
    \centering
    \includegraphics[width=0.8\columnwidth]{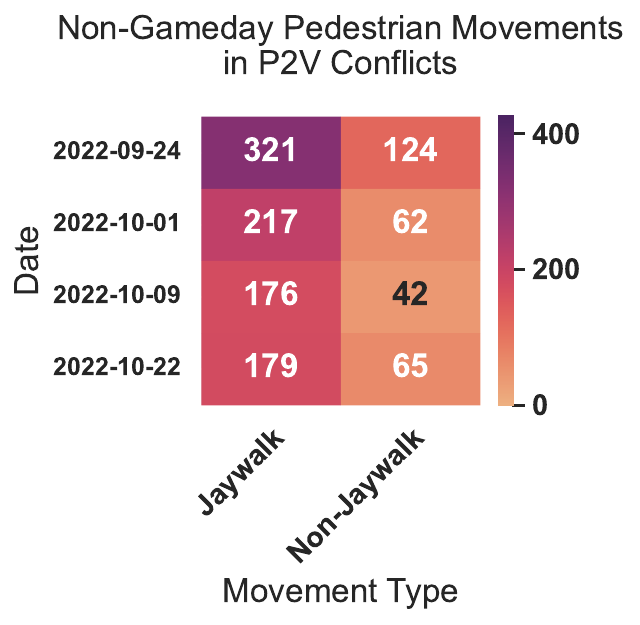}
    \captionsetup{justification=centering}
    \caption{}
    \label{fig:p2v-ped-movements-nongameday}
\end{subfigure}
\caption{Prevalence of jaywalking in P2V conflicts for both gamedays and non-gamedays.}
\label{fig:jaywalk-nonjaywalk}
\vspace{-5mm}
\end{figure*}

\subsubsection{Pedestrian-Vehicle Conflicts}

Figure~\ref{fig:p2v-normalized-conflicts} depicts the normalized counts of P2V conflicts for both gamedays and non-gamedays. On gamedays, there is a noticeable increase in the number of conflicts, particularly a few hours before the start of the games. During this period, there is also a corresponding uptick in pedestrian volume, as previously shown in Figure~\ref{fig:ped-vol-heatmaps}. As more people gather around the intersection, there is a higher likelihood of conflicts occurring between pedestrians and vehicles, especially during peak hours before the games commence.

During the non-gamedays, there is a notable increase in the number of conflicts around the afternoon hours. This trend can be attributed to pedestrians whose job shifts end in the afternoon, people visiting nearby restaurants for lunch, or socializing with friends over the weekend. The higher pedestrian activity in the vicinity during these times contributes to an elevated potential for conflicts with vehicles.

To gauge whether there is a correlation between gamedays and higher P2V conflicts, we create a lineplot that aggregates Saturdays and Sundays, both with and without football games, in Figure~\ref{fig:aggregated-p2v}. Saturday gamedays are especially frequent for P2V conflicts, with peaks occurring at 10:00 and 16:00. The figure also contains a range for ``Saturday Gamedays'' and ``Saturday Non-Gamedays'', as the dataset contains multiple days that fit these criteria (whereas the Sundays have one day each). Thus, the shaded areas around those lines represent the confidence intervals of the variation in number of conflicts during those hours of the day. The bold line represents the average number of conflicts for each hour, whereas the shaded areas represent the confidence interval. Figure~\ref{fig:aggregated-p2v} reinforces our observation of increased conflicts during gamedays on Saturdays.

% \begin{figure}[h!]
% \centering

% \begin{minipage}[b]{0.45\textwidth}
% \centering
% \includegraphics[height=7cm]{figures/5060-1-conflicts.pdf}
% \end{minipage}
% \hfill

% \begin{minipage}[b]{0.45\textwidth}
% \centering
% \includegraphics[height=7cm]{figures/5060-2-conflicts.pdf}
% \end{minipage}

% \vspace{0.5cm}

% \begin{minipage}[b]{0.45\textwidth}
% \centering
% \includegraphics[height=7cm]{figures/5060-3-conflicts.pdf}
% \end{minipage}
% \hfill
% \begin{minipage}[b]{0.45\textwidth}
% \centering
% \includegraphics[height=7cm]{figures/5060-4-conflicts.pdf}
% \end{minipage}

% \vspace{0.5cm}

% \begin{minipage}[b]{0.45\textwidth}
% \centering
% \includegraphics[height=7cm]{figures/5060-5-conflicts.pdf}
% \end{minipage}
% \hfill
% \begin{minipage}[b]{0.45\textwidth}
% \centering
% \includegraphics[height=7cm]{figures/5060-6-conflicts.pdf}
% \end{minipage}

% \caption{Heatmaps for each individual P2V type for four gamedays.}\label{fig:indiv-type-heatmaps}
% \end{figure}

Further, using our software, we categorized the P2V conflicts into different subtypes to discover patterns or commonalities or determine if a countermeasure can be taken to prevent such conflicts. The distribution of P2V conflicts over gamedays and non-gamedays is visualized in Figure~\ref{fig:conflict-type-comparison}. Conflict types 1 and 5 are the most common conflict types, where the former involves a right-turning vehicle and the pedestrian on an adjacent parallel crosswalk; the latter involves a conflict with a through vehicle and a pedestrian on the far-side crosswalk (as previously described in Figures~\ref{fig:naming-convention} and~\ref{fig:conflict-types}).

Figure~\ref{fig:indiv-type-heatmaps} shows the two prominent types of P2V conflicts for gamedays and non-gamedays: conflict types 1 and 5. The figure shows a noticeable increase in both Type 1 and 5 conflicts on gamedays, with Type 5 peaks reaching as high as 65 dangerous scenarios between pedestrians and vehicles during the late afternoon. This peak is higher than the peak of 42 during the late afternoon non-gameday.

%%%%%%

%%%%%%%%%%%%

Figure~\ref{fig:intersec-heatmap-p2v} presents a spatial heatmap of conflicts during the gamedays and non-gamedays, highlighting the areas in the intersection where the concentration of conflicts is highest. Conflict density is calculated by performing kernel density estimation (KDE) to estimate the probability density function, which is represented by the Conflict Density colorbar. Similarly, Figure~\ref{fig:histogram-heatmap-p2v} visualized the areas within the intersection where conflicts are more likely to occur regardless of their proximity to other areas. These histograms are crucial visual representations of the spatial distribution of P2V conflicts within the intersection during football game events. Notably, more dots are present in the gameday heatmaps; they also have, on average, more conflicts per day, with 507.60 conflicts on Saturday gamedays as compared to 322.67 on Saturday non-gamedays. Similarly, Sunday gamedays have 302.00 conflicts and Sunday non-gamedays have 218.00, on average.

Additionally, as seen in Figure~\ref{fig:histogram-heatmap-p2v}, some conflicts occur outside the confines of the crosswalk. This is likely due to the sheer number of pedestrians crossing, where some are forced to walk along the outer edge of the crosswalk. % Along with Figure~\ref{fig:indiv-type-heatmaps}, we can infer that a majority of the conflicts happen with right-turning vehicles.

\begin{figure*}[htp]
\centering

\begin{subfigure}[t]{.6\textwidth} % Adjust the width based on your preference
    \centering
    \includegraphics[width=1.1\textwidth]{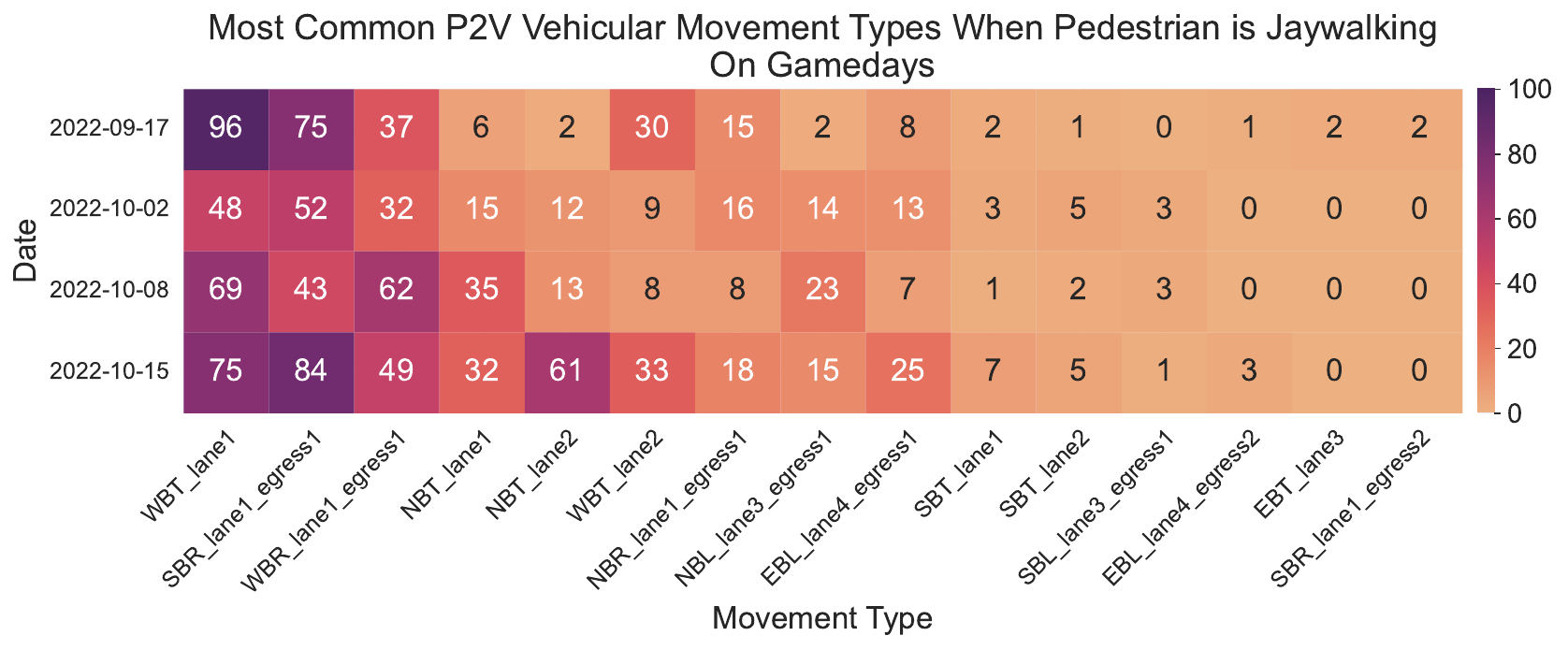}
    \captionsetup{justification=justified, singlelinecheck=false} % Adjust the justification here
    \caption{}
    \label{fig:jaywalk-gameday}
\end{subfigure}

\vspace{0.5cm} % Adjust the value for vertical space

\begin{subfigure}[t]{.6\textwidth} % Adjust the width based on your preference
    \centering
    \includegraphics[width=1.1\textwidth]{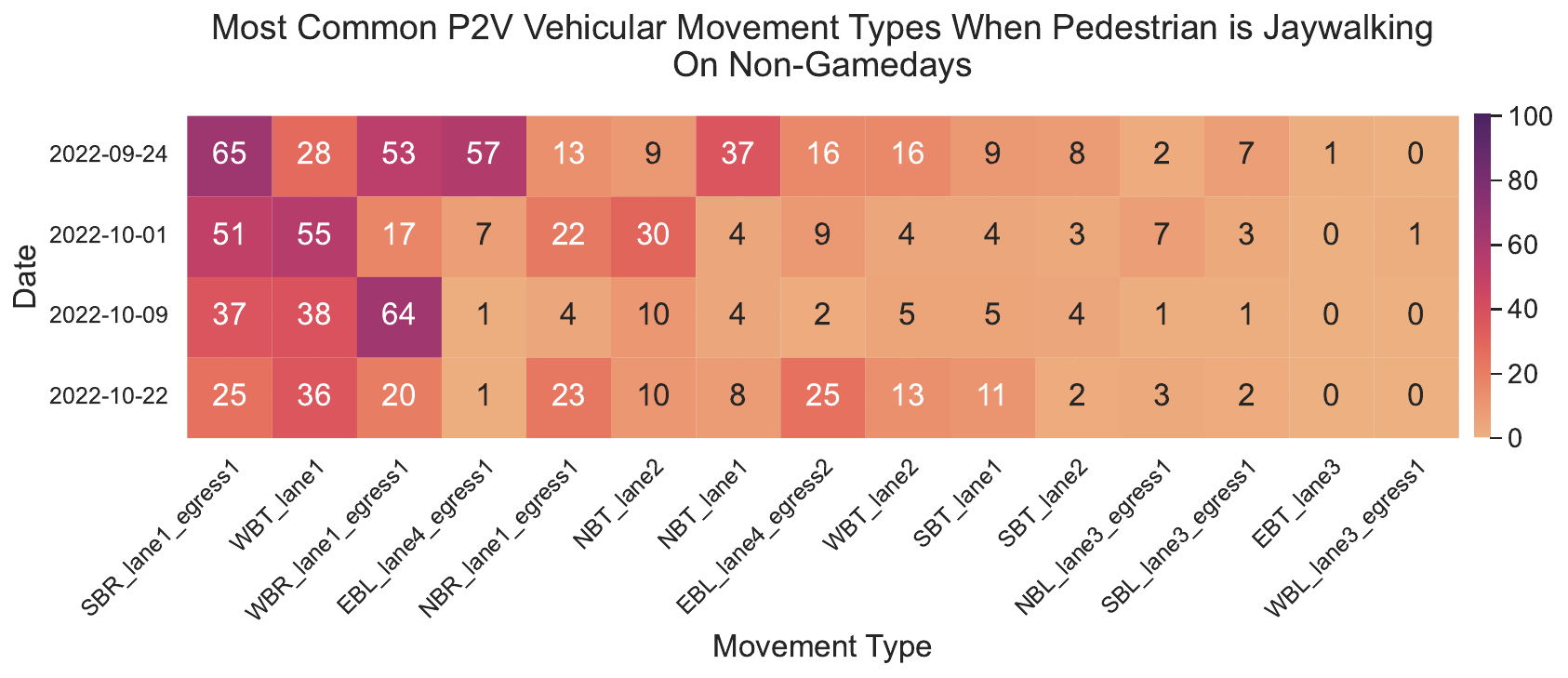}
    \captionsetup{justification=justified, singlelinecheck=false} % Adjust the justification here
    \caption{}
    \label{fig:jaywalk-nongameday}
\end{subfigure}

\caption{Movements for gamedays and non-gamedays.}
\label{fig:jaywalk-movements}
\end{figure*}

The other predominant type of conflict involves pedestrians violating the walk signal and starting to cross the crosswalk, as seen in Figure~\ref{fig:jaywalk-nonjaywalk}. Such jaywalking occurs in every P2V conflict type. To further understand how such conflicts with jaywalking pedestrians arise, Figure~\ref{fig:jaywalk-movements} shows the most common vehicle movements that are found in jaywalking pedestrian P2V conflicts. Notably, the ``WBT'', or the west-bound through, is most common on gamedays, which indicates a Type 5 conflict where the vehicle is intersecting the pedestrian's path perpendicularly. This interaction is particularly dangerous as the vehicle usually is proceeding through a green light at a high speed. On non-gamedays, the ``SBR'', or the south-bound right, is the most common vehicle movement found in jaywalking pedestrian P2V conflicts.

%By examining the heatmap, we can discern patterns and hotspots of conflicts that tend to occur during gamedays. This information is vital for traffic management and safety planning during such events. The identified conflict-prone areas can be targeted for interventions, such as improved signage, pedestrian crossings, or traffic signal adjustments, to mitigate potential risks and ensure smoother traffic flow.

% insert heatmap and describe observations
\subsubsection{Vehicle-Vehicle Conflicts}
\begin{figure*}[!ht]
\centering
\begin{subfigure}[t]{.45\textwidth}
    \centering
    \includegraphics[width=0.77\columnwidth]{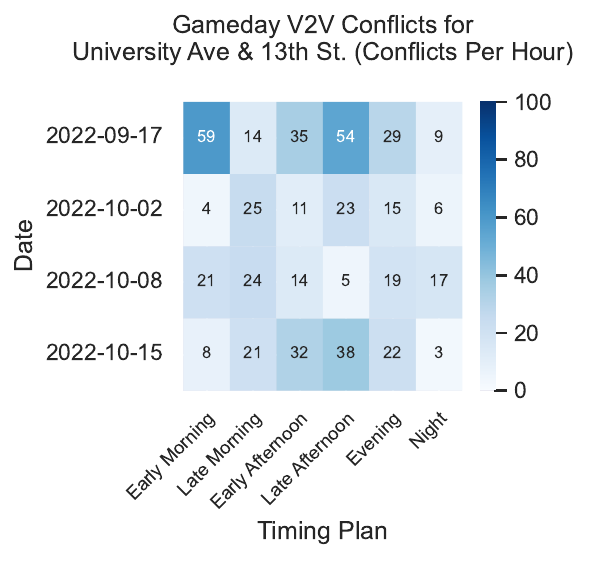}
    \captionsetup{justification=centering}
    \caption{}
    \label{fig:v2v-normalized-heatmap}
\end{subfigure}
\hspace{1cm} % Adjust the value as needed for the desired horizontal space
\begin{subfigure}[t]{.45\textwidth}
    \centering
    \includegraphics[width=0.77\columnwidth]{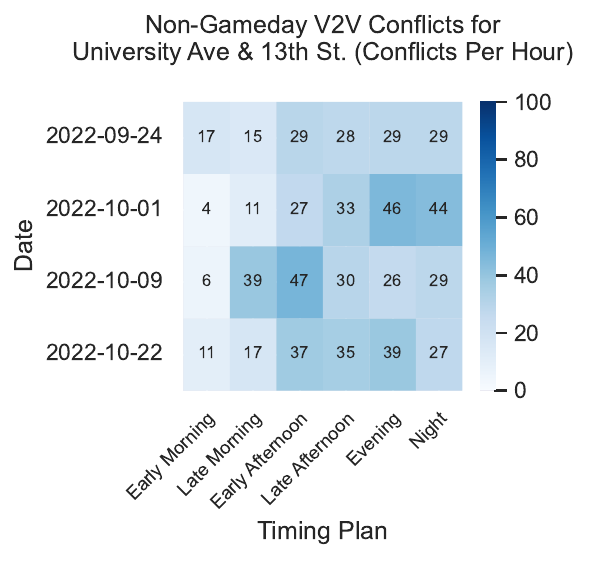}
    \captionsetup{justification=centering}
    \caption{}
    \label{fig:v2v-non-gameday-normalized-heatmap}
\end{subfigure}
\caption{Heatmaps that show the total V2V conflicts.}
\label{fig:v2v-normalized-conflicts}
\end{figure*}

Figure~\ref{fig:v2v-normalized-conflicts} illustrates the normalized counts of vehicle-to-vehicle (V2V) conflicts for both gamedays and non-gamedays. Surprisingly, on gamedays, there is a slight decrease in the number of conflicts. This could be attributed to drivers being more cautious and attentive due to the increased pedestrian presence during these events. Additionally, due to multiple road closures, vehicle speeds are lower than average. % we probably need to put a figure that proves this

On the other hand, during the non-gamedays, there is a noticeable increase in the number of conflicts compared to gamedays. The absence of football-related traffic and pedestrian congestion might result in a higher concentration of vehicles at the intersection, leading to an elevated potential for V2V conflicts.

\begin{figure}[htp]
\centering
\includegraphics[height=6cm]{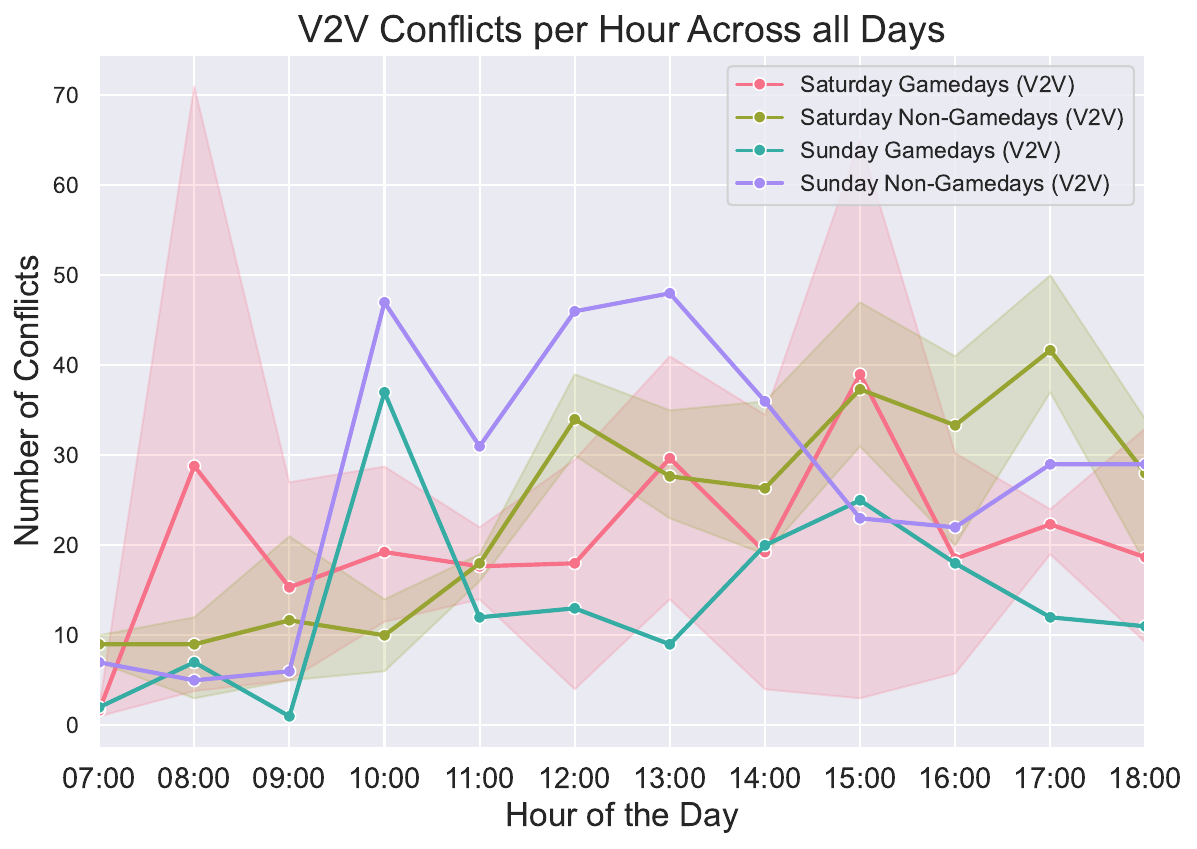}
\caption{Aggregated V2V conflicts from 07:00 to 18:00.}
\label{fig:aggregated-v2v}
\end{figure}

We show an aggregated line plot in Figure~\ref{fig:aggregated-v2v} for vehicle-to-vehicle (V2V) conflicts to find a correlation between weekday and/or gameday. This figure features similar peaks, such as the Saturday gamedays containing the highest number in the early morning (08:00) and late afternoon (15:00). However, the Sunday non-gamedays hold the highest number of V2V conflicts for most of the day (from late morning to early afternoon, inclusive). The Saturday non-gamedays also hold the highest V2V conflicts on average for the remainder of the day. Thus, gamedays tend to show fewer V2V conflicts than non-gamedays, so our analysis predominately focuses on P2V interactions on gamedays.

%insert heatmap and describe observations
%drill into particular aspects such as what type of conflict are more prominent, and why. Provide more examples, etc.

\section*{Acknowledgment}

The work was supported in part by the Florida Department of Transportation. The opinions, findings, and conclusions expressed in this publication are those of the author(s) and not necessarily those of the Florida Department of Transportation or the U.S. Department of Transportation.

The authors would like to thank Dr. Gregor von Laszewski and Ruochen Gu for their informative and invaluable discussions on data visualization, which greatly helped enhance our paper.

% \section*{References}

\bibliographystyle{IEEEtran}

\bibliography{sources.bib}

\begin{thebibliography}{10}
\providecommand{\url}[1]{#1}
\csname url@samestyle\endcsname
\providecommand{\newblock}{\relax}
\providecommand{\bibinfo}[2]{#2}
\providecommand{\BIBentrySTDinterwordspacing}{\spaceskip=0pt\relax}
\providecommand{\BIBentryALTinterwordstretchfactor}{4}
\providecommand{\BIBentryALTinterwordspacing}{\spaceskip=\fontdimen2\font plus
\BIBentryALTinterwordstretchfactor\fontdimen3\font minus \fontdimen4\font\relax}
\providecommand{\BIBforeignlanguage}[2]{{%
\expandafter\ifx\csname l@#1\endcsname\relax
\typeout{** WARNING: IEEEtran.bst: No hyphenation pattern has been}%
\typeout{** loaded for the language `#1'. Using the pattern for}%
\typeout{** the default language instead.}%
\else
\language=\csname l@#1\endcsname
\fi
#2}}
\providecommand{\BIBdecl}{\relax}
\BIBdecl

\bibitem{roundabouts}
\BIBentryALTinterwordspacing
I.~DOT, ``Roundabouts safety and efficiency,'' 2020. [Online]. Available: \url{https://iowadot.gov/traffic/roundabouts/Frequently-Asked-Questions/Roundabouts-safety-and-efficiency}
\BIBentrySTDinterwordspacing

\bibitem{isafety}
\BIBentryALTinterwordspacing
FHWA, ``Intersection safety,'' 2020. [Online]. Available: \url{https://highways.dot.gov/safety/intersection-safety/about}
\BIBentrySTDinterwordspacing

\bibitem{fars}
\BIBentryALTinterwordspacing
F.~D. of~Transportation, ``Fatality analysis reporting system (fars),'' 2020. [Online]. Available: \url{https://www-fars.nhtsa.dot.gov/Main/index.aspx}
\BIBentrySTDinterwordspacing

\bibitem{costs}
\BIBentryALTinterwordspacing
------, ``Historical item average costs reports,'' 2021. [Online]. Available: \url{https://www.fdot.gov/programmanagement/estimates/documents/historicalitemaveragecostsreports}
\BIBentrySTDinterwordspacing

\bibitem{fries}
\BIBentryALTinterwordspacing
R.~N. Fries, M.~R. Gahrooei, M.~Chowdhury, and A.~J. Conway, ``Meeting privacy challenges while advancing intelligent transportation systems,'' \emph{Transportation Research Part C: Emerging Technologies}, vol.~25, pp. 34--45, 2012. [Online]. Available: \url{https://www.sciencedirect.com/science/article/pii/S0968090X12000526}
\BIBentrySTDinterwordspacing

\bibitem{banerjee}
T.~Banerjee, K.~Chen, A.~Almaraz, R.~Sengupta, Y.~Karnati, B.~Grame, E.~Posadas, S.~Poddar, R.~Schenck, J.~Dilmore, S.~Srinivasan, A.~Rangarajan, and S.~Ranka, ``A modern intersection data analytics system for pedestrian and vehicular safety,'' in \emph{2022 IEEE 25th International Conference on Intelligent Transportation Systems (ITSC)}, 2022, pp. 3117--3124.

\bibitem{villiers}
\BIBentryALTinterwordspacing
C.~Villiers, L.~D. Nguyen, and J.~Zalewski, ``Evaluation of traffic management strategies for special events using probe data,'' \emph{Transportation Research Interdisciplinary Perspectives}, vol.~2, p. 100052, 2019. [Online]. Available: \url{https://www.sciencedirect.com/science/article/pii/S2590198219300521}
\BIBentrySTDinterwordspacing

\bibitem{yolov4}
\BIBentryALTinterwordspacing
A.~Bochkovskiy, C.~Wang, and H.~M. Liao, ``Yolov4: Optimal speed and accuracy of object detection,'' \emph{CoRR}, vol. abs/2004.10934, 2020. [Online]. Available: \url{https://arxiv.org/abs/2004.10934}
\BIBentrySTDinterwordspacing

\bibitem{kathuria}
\BIBentryALTinterwordspacing
A.~Kathuria and P.~Vedagiri, ``Evaluating pedestrian vehicle interaction dynamics at un-signalized intersections: A proactive approach for safety analysis,'' \emph{Accident Analysis \& Prevention}, vol. 134, p. 105316, 2020. [Online]. Available: \url{https://www.sciencedirect.com/science/article/pii/S0001457519303847}
\BIBentrySTDinterwordspacing

\bibitem{olszewski}
\BIBentryALTinterwordspacing
P.~Olszewski, I.~Buttler, W.~Czajewski, P.~Dąbkowski, C.~Kraśkiewicz, P.~Szagała, and A.~Zielińska, ``Pedestrian safety assessment with video analysis,'' \emph{Transportation Research Procedia}, vol.~14, pp. 2044--2053, 2016, transport Research Arena TRA2016. [Online]. Available: \url{https://www.sciencedirect.com/science/article/pii/S2352146516301739}
\BIBentrySTDinterwordspacing

\bibitem{warchol}
\BIBentryALTinterwordspacing
S.~Warchol and A.~Musunuru, ``Smart intersections: Using video analytics to identify crash patterns before they happen,'' 2023. [Online]. Available: \url{https://www.kittelson.com/ideas/smart-intersections-using-video-analytics-to-identify-crash-patterns-before-they-happen/}
\BIBentrySTDinterwordspacing

\bibitem{mishra}
\BIBentryALTinterwordspacing
A.~Mishra, K.~Chen, S.~Poddar, E.~Posadas, A.~Rangarajan, and S.~Ranka, ``Using video analytics to improve traffic intersection safety and performance,'' in \emph{Vehicles}, vol. 4(4), 2022, pp. 1288--1313. [Online]. Available: \url{https://www.mdpi.com/2624-8921/4/4/68}
\BIBentrySTDinterwordspacing

\bibitem{austin}
H.~Austin, A.~Freed, A.~Labus, B.~Lynch, J.~Mastrullo, J.~Sharff, and R.~J. Riggs, ``A systems approach to improving the spectator experience at collegiate football games,'' in \emph{2023 Systems and Information Engineering Design Symposium (SIEDS)}, 2023, pp. 214--219.

\bibitem{xiong}
G.~Xiong, X.~Dong, D.~Fan, F.~Zhu, K.~Wang, and Y.~Lv, ``Parallel traffic management system and its application to the 2010 asian games,'' \emph{IEEE Transactions on Intelligent Transportation Systems}, vol.~14, no.~1, pp. 225--235, 2013.

\bibitem{kattan}
\BIBentryALTinterwordspacing
L.~Kattan, S.~Acharjee, and R.~Tay, ``Pedestrian scramble operations: Pilot study in calgary, alberta, canada,'' \emph{Transportation Research Record}, vol. 2140, no.~1, pp. 79--84, 2009. [Online]. Available: \url{https://doi.org/10.3141/2140-08}
\BIBentrySTDinterwordspacing

\bibitem{huang}
X.~Huang, T.~Banerjee, K.~Chen, N.~Varanasi, A.~Rangarajan, and S.~Ranka, ``Machine learning based video processing for real-time near-miss detection,'' in \emph{6th International Conference on Vehicle Technology and Intelligent Transport Systems (VEHITS)}, 01 2020, pp. 169--179.

\bibitem{tania-book}
\BIBentryALTinterwordspacing
T.~Banerjee, X.~Huang, A.~Wu, K.~Chen, A.~Rangarajan, and S.~Ranka, \emph{{Video Based Machine Learning for Traffic Intersections}}.\hskip 1em plus 0.5em minus 0.4em\relax Philadelphia, PA, USA: Taylor {\&} Francis Group, Oct. 2023. [Online]. Available: \url{https://books.google.com/books/about/VIDEO\_BASED\_MACHINE\_LEARNING\_FOR\_TRAFFIC.html?id=28HlzwEACAAJ}
\BIBentrySTDinterwordspacing

\bibitem{OpenStreetMap}
{OpenStreetMap contributors}, ``{Planet dump retrieved from https://planet.osm.org },'' \url{ https://www.openstreetmap.org }, 2023.

\bibitem{aadt-report}
\BIBentryALTinterwordspacing
``{2022 Annual Average Daily Traffic Report},'' Mar. 2023, [Online; accessed 27. Jul. 2023]. [Online]. Available: \url{https://tdaappsprod.dot.state.fl.us/fto/reports/622UPD\_Combined\_AADT\_Report\_2022/2\_26\_CAADT.pdf}
\BIBentrySTDinterwordspacing

\bibitem{henderson}
\BIBentryALTinterwordspacing
J.~Henderson, ``{Pedestrians and bicyclists keep getting hit by cars in Gainesville. What's being done?}'' \emph{Gainesville Sun}, Apr. 2022. [Online]. Available: \url{https://www.gainesville.com/story/news/2022/04/08/people-keep-getting-hit-cars-gainesville-whats-being-done/9494756002}
\BIBentrySTDinterwordspacing

\bibitem{gnv-traffic-crashes}
\BIBentryALTinterwordspacing
``{statGNV - Traffic Crashes},'' Jan. 2023, [Online; accessed 27. Jul. 2023]. [Online]. Available: \url{https://stat.cityofgainesville.org/crashes.html}
\BIBentrySTDinterwordspacing

\bibitem{cfbd}
\BIBentryALTinterwordspacing
B.~Radjewski, ``{About},'' \emph{CFBD Blog}, Feb. 2020. [Online]. Available: \url{https://blog.collegefootballdata.com/about}
\BIBentrySTDinterwordspacing

\bibitem{parks}
\BIBentryALTinterwordspacing
J.~Parks, ``{How long are college football games? Here's what you need to know},'' \emph{College Football HQ}, Oct. 2022. [Online]. Available: \url{https://www.si.com/fannation/college/cfb-hq/ncaa-football/how-long-are-college-football-games-what-you-need-to-know}
\BIBentrySTDinterwordspacing

\bibitem{smits}
\BIBentryALTinterwordspacing
G.~Smits, ``{It's about time: College football steadily moving to a four-hour experience},'' \emph{Florida Times-Union}, Aug. 2017. [Online]. Available: \url{https://www.jacksonville.com/story/sports/college/fsu-seminoles/2017/08/05/it-s-about-time-college-football-steadily/985480007}
\BIBentrySTDinterwordspacing

\end{thebibliography}

\end{document}